\documentclass[lettersize,journal]{IEEEtran}
\usepackage{amsmath,amsfonts}
\usepackage{algorithmic}
\usepackage{algorithm}
\usepackage{array}
\usepackage[caption=false,font=normalsize,labelfont=sf,textfont=sf]{subfig} %
\usepackage{textcomp}
\usepackage{stfloats}
\usepackage{url}
\usepackage{verbatim}
\usepackage{graphicx}
\usepackage{cite}
\usepackage[bookmarks=true,pdfencoding=auto,breaklinks,colorlinks,hypertexnames=false]{hyperref} %
\usepackage{amssymb}
\usepackage{xcolor}
\usepackage{dsfont}
\usepackage{siunitx}
\usepackage{colortbl}
\usepackage{booktabs}
\usepackage{multirow}
\usepackage{makecell}
\usepackage{microtype}
\usepackage{cuted}
\usepackage{bibunits}
\usepackage{pifont}
\usepackage{flushend}

\usepackage{cleveref}

\definecolor{bestgreen}{HTML}{74AA9C}
\definecolor{secondyellow}{HTML}{FAC565}
\definecolor{thirdblue}{HTML}{83bbf1}

\newcommand{\bestcell}{\cellcolor{bestgreen!40}}
\newcommand{\secondcell}{\cellcolor{secondyellow!40}}
\newcommand{\thirdcell}{\cellcolor{thirdblue!40}}

\newcommand{\besttext}[1]{\colorbox{bestgreen!40}{#1}}
\newcommand{\secondtext}[1]{\colorbox{secondyellow!40}{#1}}
\newcommand{\thirdtext}[1]{\colorbox{thirdblue!40}{#1}}

\usepackage{xspace}
\newcommand{\methodname}{SCOUT\xspace}
\newcommand{\methodfullname}{Scene Graph-Based Exploration with Learned Utility for Open-World Interactive Object Search\xspace}
\newcommand{\benchmark}{SymSearch\xspace}

\begin{document}

\title{Relational Semantic Reasoning on 3D Scene Graphs \\ for Open World Interactive Object Search}

\author{
Imen Mahdi\textsuperscript{1},\quad Matteo Cassinelli\textsuperscript{2},\quad Fabien Despinoy\textsuperscript{2},\quad Tim Welschehold\textsuperscript{1},\quad Abhinav Valada\textsuperscript{1}
\vspace{0.3cm}
\newline
\textsuperscript{1}University of Freiburg\quad\quad\textsuperscript{2}Toyota Motor Europe
}

\maketitle

\begin{abstract}
Open-world interactive object search in household environments requires understanding semantic relationships between objects and their surrounding context to guide exploration efficiently. Prior methods either rely on vision–language embeddings similarity, which does not reliably capture task-relevant relational semantics, or large language models~(LLMs), which are too slow and expensive for real-time deployment. We introduce \methodname: \underline{SC}ene Graph-Based Expl\underline{O}ration with Learned \underline{U}tility for Open-World In\underline{T}eractive Object Search, a novel method that searches directly over 3D scene graphs by assigning utility scores to rooms, frontiers, and objects using relational exploration heuristics such as room--object containment and object--object co-occurrence. To make this practical without sacrificing open-vocabulary generalization, we propose an offline procedural distillation framework that extracts structured relational knowledge from LLMs into lightweight models for on-robot inference. Furthermore, we present \benchmark, a scalable symbolic benchmark for evaluating semantic reasoning in interactive object search tasks. Extensive evaluations across symbolic and simulation environments show that \methodname outperforms embedding similarity-based methods and matches LLM-level performance while remaining computationally efficient. Finally, real-world experiments demonstrate effective transfer to physical environments, enabling open-world interactive object search under realistic sensing and navigation constraints.
\end{abstract}


\IEEEpeerreviewmaketitle

\begin{figure*}[t]
    \centering
    \includegraphics[width=\linewidth]{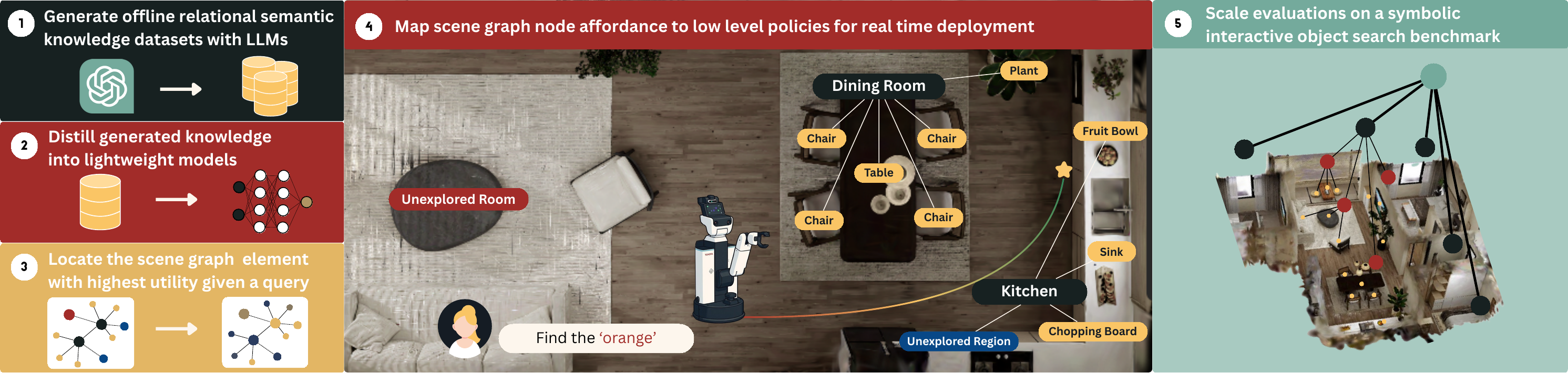}
    \caption{\textbf{Overview of \methodname: \methodfullname.}
    \methodname procedurally distills structured, relational semantic knowledge between scene elements from large language models into lightweight models (1–2). During exploration, the agent assigns utility scores to scene graph nodes based on exploration heuristics previously learned (3) and grounds high-level actions through low-level navigation and manipulation policies (4). \benchmark, our symbolic benchmark (5), enables scalable evaluation of relational semantic reasoning over scene graphs with no simulation overhead.}
    \label{fig:cover}
\end{figure*}
\section{Introduction}
\label{sec:intro}

\IEEEPARstart{R}{obots} operating in household environments must be able to efficiently search for objects, even when the target is hidden inside containers. Humans solve such tasks by leveraging rich semantic priors, such as what objects typically co-occur with and which rooms commonly contain them, which help guide the search without exhaustive exploration~\cite{hidalgo2005human,mack2011object}. Endowing robots with similar relational knowledge remains challenging, especially in open-world settings where queries can reference arbitrary object categories.

Most existing object-search methods implicitly approximate these semantic priors using similarity between vision–language embeddings and a target query \cite{gadre2023cows,yokoyama2024vlfm,bajpai2025uncertainty,zhang2025apexnav}. While effective for measuring visual or functional similarity, such embeddings do not reliably encode relational semantics. For instance, “milk carton” may appear similar to “fridge” and “oven” in the embedding space, even though only one is a plausible container. As a result, current embedding similarity-based search often struggles to distinguish semantically meaningful locations. This problem becomes more apparent when objects require interaction with the environment to be revealed, e.g, opening containers. Another class of methods \cite{shah2023navigation,wu2024voronav,loo2025open,yin2024sg,ge2024commonsense,zhang2025language,menonopen,chalvatzaki2023learning,liu2404delta,booker2024embodiedrag,honerkamp2024language,mohammadi2025more} relies on pretrained large language models (LLMs) to implicitly incorporate these semantic priors into their search pipelines. LLMs encode commonsense knowledge on human-centered environments and support open-vocabulary queries. However, online planning with LLMs is computationally expensive and unrealistic for autonomous agents with hardware constraints, as repeated queries scale poorly with scene complexity \cite{parashar2025inference}. Knowledge distillation approaches \cite{ha2023scaling,ravichandran2025distilling} mitigate this issue but require complex data-generation pipelines and long training times.

In parallel, 3D scene graphs (3DSGs) \cite{armeni20193d,werby2024hierarchical,steinke2025collaborative} have emerged as compact, semantically structured representations for scene understanding. Their hierarchical structure makes them well-suited for relational reasoning tasks. However, most existing approaches utilize them to ground LLM-based planners~\cite{rana2023sayplan,liu2404delta,booker2024embodiedrag,honerkamp2024language,mohammadi2025more,loo2025open,yin2024sg}, leaving open the question of how to efficiently exploit their embedded semantics without incurring high inference-time costs.

In this work, we propose \methodname: \methodfullname, an interactive object-search method that performs reasoning directly at the scene graph level. SCOUT assigns utility scores to rooms, regions, objects, and containers for a given query using two key exploration heuristics: room–object containment and object–object co-occurrence. These scores guide exploration toward the most promising scene elements without requiring computationally expensive online LLM calls. To efficiently estimate these utility scores, we introduce an offline procedural distillation framework that extracts structured relational semantic knowledge from LLMs and trains lightweight models to predict containment and co-occurrence priors over scene elements. These distilled models retain open-vocabulary generalization while enabling on-robot inference. Finally, we present \benchmark, a symbolic, scene-graph-based benchmark that evaluates open-vocabulary relational semantic reasoning without simulation overhead. \benchmark realistically simulates how an embodied agent incrementally discovers rooms, frontiers, and objects, enabling scalable evaluation of exploration strategies. We provide an overview of our work in Fig.~\ref{fig:cover}.

In summary, our main contributions are the following:
\begin{enumerate}
    \item \methodname, a heuristics-based exploration method operating directly on 3D scene graphs.
    \item A procedural LLM distillation framework that captures open-vocabulary relational semantic knowledge for real-time inference.
    \item \benchmark, a scalable symbolic benchmark for open-world interactive object search.
    \item A quantitative analysis of embedding limitations and a comprehensive evaluation showing that \methodname surpasses embedding-based baselines and approaches LLM-level reasoning at a fraction of the computational cost, with successful transfer to real-world robots.
    \item We make the code publicly available at \url{https://scout.cs.uni-freiburg.de}.
\end{enumerate}
\section{Related Work}
\label{sec:related}
Object search, also referred to as object goal navigation, in unknown environments has been widely studied~\cite{schmalstieg2023learning,schmalstieg2022learning}. While classical approaches typically assume a closed set of object categories, realistic object search requires handling open-vocabulary queries. Recent methods address this by leveraging vision--language embeddings to estimate the relevance of observed regions to a textual query, using similarity scores as search heuristics. Common strategies for these methods include constructing relevance or semantic maps to score frontiers~\cite{gadre2023cows,yokoyama2024vlfm,bajpai2025uncertainty,zhang2025apexnav,prasanna2024perception} or learning semantic-driven navigation policies~\cite{gadre2023cows}. Nevertheless, embedding similarity often yields noisy or ambiguous signals.

Alternatively, large language models have recently been adopted in robotic planning tasks due to their ability to reason over long horizons, handle open-vocabulary queries, and encode commonsense knowledge~\cite{chisari2025robotic}. Among these methods~\cite{shah2023navigation,wu2024voronav} use them to score frontiers and spatial regions. However, these approaches assume that the target object can be observed without interaction. A central challenge in interactive exploration with LLMs is grounding reasoning in the physical world. Many approaches rely on 3D scene graphs~\cite{armeni20193d} as compact and semantically rich representations that can be easily presented in textual format and constrain LLM outputs to a predefined set of high-level actions with arguments grounded in scene graph elements~\cite{rana2023sayplan,chalvatzaki2023learning,liu2404delta,booker2024embodiedrag,mohammadi2025more}.

Several recent works apply LLMs and 3DSGs specifically to open-world object search by using them to select regions and objects to explore~\cite{loo2025open,yin2024sg} or to assign explicit relevance scores to candidate receptacles based on object and room categories~\cite{menonopen}. MoMa-LLM~\cite{honerkamp2024language} addresses interactive exploration by exposing a skill API that grounds abstract manipulation and exploration actions to elements of the 3DSG. To reduce LLM calls, GODHS~\cite{zhang2025language} first exhaustively explores all rooms, then uses an LLM to rank rooms by their likelihood of containing the target object, and finally identifies and ranks receptacles within each room. Closer to our work is Seek~\cite{ginting2024seek} that trains relational semantic networks on small distilled data from LLMs and 3DSGs to guide exploration in industrial inspection tasks. We address interactive object search by explicitly incorporating search heuristics over scene graphs. We perform the search directly at the scene graph level by selecting the scene element with the highest utility to explore and mapping it to low-level policies via node affordances. To estimate utility in an open-vocabulary setting, we procedurally extract large relational semantic knowledge offline using LLMs and construct supervised datasets to train lightweight models suitable for real-time deployment.

Existing object search benchmarks~\cite{savva2019habitat,szot2021habitat,yadav2023habitat,yokoyama2024hm3d} primarily evaluate perception and navigation, providing limited support for semantic reasoning or interactive environments. Existing interactive simulators include AI2-THOR~\cite{kolve2017ai2}, which spans one-room scenes, and OmniGibson~\cite{li2024behavior}, which is computationally expensive and often unstable for generating scenes. Taskography~\cite{agia2022taskography} is a symbolic planning benchmark that enables fast evaluation using scene graphs but focuses on simple rearrangement tasks and does not evaluate semantic reasoning. In contrast, we propose a new symbolic benchmark designed to scale semantic reasoning evaluation over large, open-vocabulary 3D scene graphs that more closely resemble real household environments and offer substantially greater semantic diversity and scene complexity.
\label{sec:sgbe}
\begin{figure*}
    \centering
    \includegraphics[width=\textwidth]{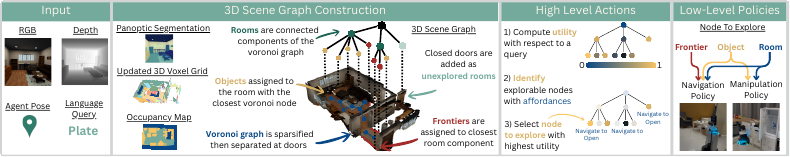}
    \caption{\textbf{Full pipeline of \methodname illustrated}. From left to right: 3DSG is constructed online from raw observations. Scene graph nodes are scored based on their utility in finding the query. Once the node to explore is selected, its affordances determine which low-level policies to run.}
    \label{fig:pipeline}
\end{figure*}

\section{Problem Formulation}
\label{sec:problem}
We study \emph{open-world interactive object search} in unknown household environments. The agent receives an open-vocabulary language query $q$ describing a target object and must locate it by actively exploring the scene (e.g., opening cabinets or doors) as target objects may be occluded within closed containers. 
At each timestep $t$, the agent constructs an incremental 3DSG $\mathcal{G}_t$ from raw observations, such as egocentric RGB-D images $\mathcal{I}_t$ and pose estimates $P_t$ obtained from onboard odometry. 
Following~\cite{armeni20193d}, we define a hierarchical 3DSG as a directed tree $
\mathcal{G}=(\mathcal{V},\mathcal{E})$,
with $\mathcal{V}$ nodes, $\mathcal{E}$ edges, and $L$ layers of abstraction. Nodes represent scene elements such as rooms, regions, objects, or nested objects. Edges include inter-layer relations, e.g., rooms contain regions, regions contain objects, and intra-layer relations, e.g., rooms connected via doors. Each node has a unique parent in the preceding layer. Nodes carry open sets of attributes such as labels, positions, states, and affordances.
We assume that the 3DSG construction can be performed by any of the existing methods~\cite{rosinol2021kimera,hughes2022hydra,werby2024hierarchical,gu2024conceptgraphs,maggio2024clio}. In this setup, we distinguish between two levels of actions: high-level and low-level actions. High-level actions correspond to node-level affordances, e.g, navigate to and open an object, which are then executed through low-level navigation or manipulation policies. The task is considered solved when the target becomes visible, and the agent reaches it within a predefined distance threshold.
In this work, we use a 5-layer hierarchy: \emph{root}, \emph{rooms}, \emph{regions/frontiers}, \emph{objects/containers}, and \emph{nested objects}. Our goal is to leverage this structured representation to guide interactive search under open-vocabulary queries.

\section{\methodname: Scene Graph-Based Exploration with Learned Utility}
We introduce \methodname, a method that leverages the semantically rich scene graph structure, estimates how informative observed scene elements are for locating a given open-vocabulary query, and maps high-level actions to low-level policies accordingly. We first describe how 3DSGs are constructed from raw observations. We then define a \textit{utility} scoring function and describe how it is computed using exploration heuristics, and detail how these heuristics are learned via offline structured relational semantic knowledge distillation from LLMs. Finally, we detail how node-level actions are grounded by mapping affordances to low-level policies. The full pipeline is illustrated in Fig.~\ref{fig:pipeline}.

\subsection{3DSG Construction}
\label{sec:sg_construction}
Real-world environments provide fine-grained observations and require the execution of low-level actions. Following the approach of~\cite{honerkamp2024language,mohammadi2025more}, we construct the 3DSG from RGB-D observations and pose estimates. We perform semantic and instance segmentation on the RGB images and fuse the results into a semantic voxel map using depth and pose information. This voxel map is then projected into a bird’s-eye-view (BEV) semantic map and an occupancy map.
Similar to Hydra~\cite{hughes2022hydra}, we build a Voronoi-based navigation graph from the occupancy map. To identify rooms, we sparsify the navigation graph and remove edges that intersect with detected doors, resulting in connected components corresponding to individual room segments. Object instances extracted from the semantic voxel map are assigned to rooms by selecting the component associated with the nearest Voronoi node, and room categories are inferred from the objects they contain. Frontiers are extracted from the BEV occupancy map as boundaries between explored traversable space and unexplored regions~\cite{topiwala2018frontier}. Additionally, detected closed doors are treated as entrances to unexplored rooms and are added as unexplored room nodes in the graph.

\subsection{Estimating Utility via Exploration Heuristics}
In interactive object search, relational semantics between the target query and elements observed in the scene provide informative cues for deciding where to explore next. 
Prior work often relies on vision–language embedding similarity to identify promising regions~\cite{gadre2023cows,yokoyama2024vlfm,bajpai2025uncertainty,zhang2025apexnav}. However, embeddings typically capture visual or functional similarity, which is not always the most informative signal for exploration. Indoor objects exhibit strong contextual regularities: many objects co-occur with others, e.g., forks with plates and dishwashers, and follow a probability distribution over room categories, e.g., forks appearing predominantly in kitchens or dining rooms. These relational priors between a query and observed scene elements provide stronger cues for object search than visual similarity alone. 
Accordingly, we define a utility scoring function: 
\begin{equation}
\label{eq:p_cooccur}
u_q(n) \in [0,1],
\end{equation}
which captures how informative a scene-graph node $n$ is for localizing the queried object $q$.
Motivated by human-inspired search heuristics, we approximate utility using object–object co-occurrence and room–object containment probabilities. 
This utility score is assigned differently based on the node type:

\paragraph{Rooms} If a room $r$ is observed and categorized, its utility is given by
\begin{equation}
    \label{eq:p_contain}
    u_q(r) \approx p_{\mathrm{contains}}(r, q),
\end{equation}
where $p_{\mathrm{contains}}(r,q)$ is the probability that room $r$ contains the target object $q$. Unobserved rooms, e.g., detected from closed doors, are unclassified and therefore cannot be scored. To ensure they remain selectable when all visible nodes are uninformative, we assign them a default exploration score.

\paragraph{Objects and Containers}
Objects' utility is estimated through co-occurrence probabilities:
\begin{equation}
    \label{eq:p_cooccur}
    u_q(o) \approx p_{\mathrm{co\mbox{-}occur}}(o, q),
\end{equation}
where $p_{\mathrm{co\mbox{-}occur}}(o,q)$ denotes the probability that object $o$ co-occurs with target object $q$, i.e., the probability that they appear in the same location. Because an object's utility depends strongly on the room context (e.g., ``kitchen cabinet'' vs.\ ``bathroom cabinet''), we update object scores as follows:
\begin{equation}
\label{eq:updated_scores}
    u^{\mathrm{updated}}_q(o)
    = u_q(r_o)\,\bigl(w + (1-w)\,u_q(o)\bigr),
\end{equation}
where $r_o$ is the parent room of object $o$ and $w\in[0,1]$ controls how much influence the room has on the final score.

\paragraph{Frontiers}
Each frontier is assigned an aggregated score based on nearby objects within a spatial radius. If no informative objects are nearby, e.g., a hallway frontier, we instead assign the room score if the frontier lies entirely within the room, or a default exploration score if it leads outside the room, to encourage expansion into unknown space.
\subsection{Learning Relational Semantics via Structured Knowledge Distillation}
\label{sec:knowledge_distillation}
Relational semantic priors, such as co-occurrence or containment, are challenging to define for an open-set vocabulary. Typically, these heuristics are implicitly followed through online LLM planning. Instead, we extract them offline via structured querying of a pretrained LLM and distill this knowledge into lightweight supervised models that can run fast inference with minimum hardware requirements. Specifically, our goal is to construct a supervised learning dataset given a semantic relation, e.g., co-occurrence or containment, by procedurally querying an LLM
$$
\mathcal{D}_{\mathrm{relation}}
= \{(x_i, y_i)\}_{i=1}^N,
$$
where $x_i$ denotes a pair of textual inputs and $y_i$ is the corresponding utility score.

Our objective is to construct a large and diverse open-vocabulary set of household objects. A key challenge in generating large object lists with LLMs is encountering hallucinations or token limits. Moreover, repeatedly using the same prompt to generate smaller sets at each time often
results in highly overlapping object sets. To address these issues, we propose a structured, multi-stage procedural generation pipeline to construct our initial open vocabulary object set $\mathcal{O}_{\mathrm{household}}$.

For this, we use the following hierarchy:
\begin{equation*}
    \text{Household} \;\to\; \text{Rooms} \;\to\; \text{Categories} \;\to\; \text{Objects}.
\end{equation*}

More precisely, we begin by generating a set of household room categories, denoted as $\mathcal{R}$. Since the number of room types is relatively small, this step can be performed using a single prompt. Given the resulting set of rooms, we then generate, for each room category $r$, a list of associated object categories $\mathcal{C}(r)$. Finally, for each room–object category pair $(r,c)$, we generate a list of associated objects $\mathcal{O}(r,c)$. This yields
\begin{equation}
    \mathcal{O}_{\mathrm{household}}
    = \bigcup_{r\in\mathcal{R}} \ \bigcup_{c \in \mathcal{C}(r)} \mathcal{O}(r,c).
\end{equation}

All generated outputs are post-processed by converting text to lowercase, enforcing space-separated words, and normalizing entries to singular nouns. This normalization avoids treating semantically identical objects as distinct due to minor formatting variations. To further increase diversity, each prompt can be executed multiple times. The final object set is obtained by aggregating all unique outputs across runs. We manually inspected and validated subsets of the final dataset and did not observe any inconsistencies. We deliberately avoid manually curating the outputs, as our goal is to evaluate the automated data generation pipeline. The procedure is designed to be reusable for different target environments (e.g., schools, hospitals) with minimum intervention.

\begin{figure}[t]
    \centering
    \includegraphics[width=0.9\linewidth]{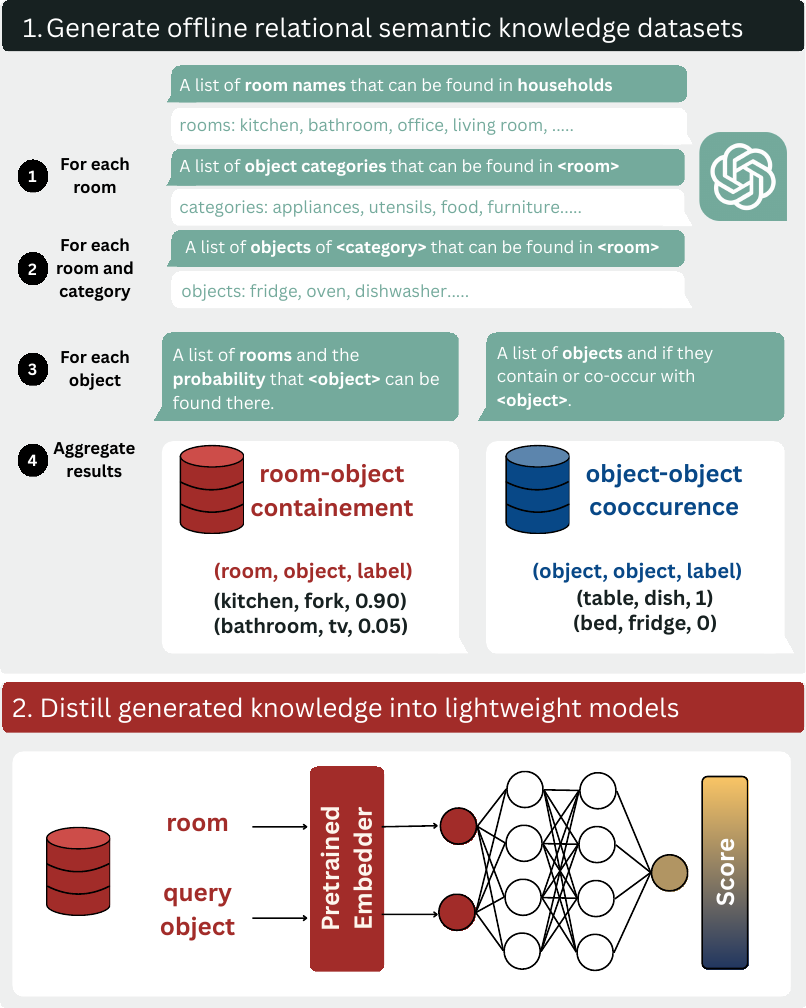}
    \caption{\textbf{Overview of the procedural knowledge distillation pipeline.}
    An LLM is queried procedurally to generate a diverse open-vocabulary set of household objects and relational annotations.
    The resulting object–object co-occurrence and room–object containment datasets are then used to supervise lightweight MLP models for utility estimation.}
    \label{fig:knowledge_distillation}
\end{figure}

Once the complete household object set is constructed, we use it to generate two relational semantic datasets. For object–object co-occurrence, the LLM proposes, for each query $q \in \mathcal{O}_{\mathrm{household}}$, a set of objects annotated with binary co-occurrence labels, i.e., 1 if the object is likely to co-occur with the target query and 0 otherwise. For room–object containment, the LLM provides a list of rooms with continuous scores $y\in[0,1]$, i.e., the probability that an object $q$ is found in room $r$. The full pipeline used to generate these datasets is shown in Fig.~\ref{fig:knowledge_distillation}.

For object co-occurrence, we experimented with both binary and continuous labels and found that binary scores are better suited for this task. We attribute this to the inherent ambiguity of object–object relationships, which are often difficult to quantify precisely (e.g., how much more likely a pen is to co-occur with a desk than with paper). In contrast, room–object associations are more structured and can be meaningfully expressed as continuous probabilities (e.g., a cup’s most probable room locations might be: kitchen, living room, bedroom, bathroom, in decreasing order).

The final datasets obtained are thus:
\begin{equation}
    \mathcal{D}_{\mathrm{co\mbox{-}occur}}=\{((o,q)_i, y_i)\}_{i=1}^N,
\end{equation}
and
\begin{equation}
    \mathcal{D}_{\mathrm{contain}} = \{((r,q)_i, y_i)\}_{i=1}^M,
\end{equation}
with $N$ and $M$ denoting the sizes of each dataset, respectively.

Once the datasets are constructed, they serve as the ground truth for training our lightweight models. To ensure open-vocabulary generalization, we first encode textual inputs using a frozen pretrained text encoder \( E(\cdot) \). Concretely, each data point consists of two text elements: a query \( q \) and a scene graph node \( n \) (representing either an object or a room). These are embedded independently and concatenated to form the input vector:
\begin{equation}
    \label{eq:vectorize}
    x = E(n) \oplus E(q), \quad x \in \mathbb{R}^{2d}
\end{equation}
where \( d \) is the embedding dimension and $\oplus$ denotes dimension-wise concatenation.

A simple parametric model
\begin{equation}
    f_\theta^{\mathrm{relation}} : \mathbb{R}^{2d} \to [0,1]
\end{equation}
is then learned via supervised learning.

Concretely, we train two MLPs: $f_{\theta_1}^{\mathrm{co\mbox{-}occur}}$ with binary cross entropy as loss function and $f_{\theta_2}^{\mathrm{contain}}$ with mean squared error as loss function. This distillation procedure ensures (i) low inference costs during online exploration and (ii) generalization to an open vocabulary set of objects and query descriptions that come from the pretrained embeddings.

Finally, these models define the utility scoring functions used to score scene graph nodes following Eq.~\ref{eq:p_contain} and~\ref{eq:p_cooccur}:
\begin{equation}
    \label{eq:score_cooccur}
    u_q(o)
    = f_{\theta_1}^{\mathrm{co\mbox{-}occur}}(E(o)\oplus E(q)),
\end{equation}
\begin{equation}
    \label{eq:score_contain}
    u_q(r)
    = f_{\theta_2}^{\mathrm{contain}}(E(r)\oplus E(q)),
\end{equation}
.

An overview of the full pipeline is shown in Fig.~\ref{fig:knowledge_distillation}.

\subsection{Selecting and Grounding High Level Actions}
After all nodes are scored, the agent considers the set of actionable nodes $\mathcal{N}_a$, i.e., nodes with exploration affordances such as openable containers, closed doors, or unexplored frontiers. To avoid excessive travel, distance is incorporated into the selection. Specifically, let $U_{\max}$ denote the maximum utility scored among actionable nodes. The agent first selects nodes with a utility score in the range $[U_{\max}-\Delta,\,U_{\max}]$, where $\Delta$ is a pre-defined margin that ensures elements with close scores are equally important, and then chooses the one with minimum travel distance:

\begin{equation}
\label{eq:distance}
    n^* = \arg\min_{n \in \mathcal{N}_a'} d(n),
\end{equation}
where $\mathcal{N}_a'=\{n \in \mathcal{N}_a \mid u_q(n)\ge U_{\max}-\Delta\}$, and $d(n)$ is the estimated geodesic distance from the current robot position to the target node $n$. Note that setting $\Delta=0$ is the same as taking the highest scoring node at each timestep greedily:
\begin{equation}
    n^* = \arg\max_{n\in\mathcal{N}_a} u_q(n).
\end{equation}

Once a scene-graph node $n^*$ is selected for exploration, its affordance dictates the subsequent action and low-level policy. For navigable nodes, e.g., frontiers or objects, the agent plans a collision-free path to the nearest Voronoi node using A* \cite{hart1968formal} on an inflated occupancy map. If a selected object is openable, the agent invokes a manipulation policy to open it. In practice, the agent first navigates to the target; if the target is a container, i.e., has an ``openable'' affordance, the agent then transitions to the manipulation policy to explore its contents.
\begin{figure*}[t]
\centering
\includegraphics[width=\linewidth]{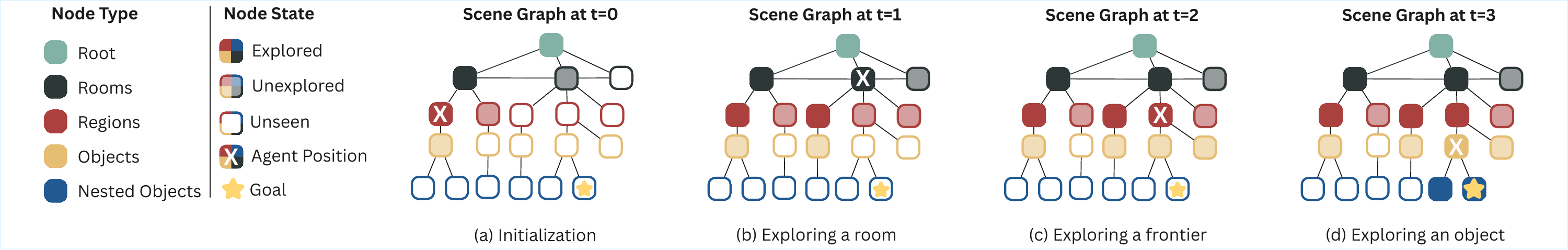}
\caption{
\textbf{Illustration of \benchmark's roll-out process.} At each step, the agent receives a scene-graph observation and a target object query. 
(a) At initialization, the agent spawns in a random region and room. Unexplored regions appear as frontier nodes, and rooms connected via doors are seen but unexplored. 
(b) Exploring a new room reveals the closest region to the agent in that room, along with its objects. The remaining regions are considered unexplored frontiers. 
(c) Exploring a frontier reveals the objects within that region. 
(d) Exploring an object reveals its nested objects (e.g., objects inside or on top of it). The episode terminates when the target object becomes visible. 
}
\label{fig:rollout}
\end{figure*}

\section{Symbolic Object Search on 3D Scene Graphs}
\label{sec:symsearch}
We propose a flexible symbolic benchmark for interactive object search that satisfies the following requirements:
\begin{itemize}
    \item \textbf{Scene-graph based representation:} uses 3D Scene Graphs, enabling compatibility with any dataset source and supporting arbitrary hierarchical depths, node attributes, and edge types.
    \item \textbf{Open-vocabulary semantics:} supports open set object and room categories and attributes.
    \item \textbf{Embodied Exploration Realism:} mimics how an embodied agent perceives and incrementally explores an environment.
    \item \textbf{Evaluation of semantic reasoning:} Focuses on evaluating high-level decision-making based on semantics.
\end{itemize}

\subsection{Dataset-Agnostic Evaluation Framework}
\begin{table*}[t]
\centering
\footnotesize
\caption{\textbf{Statistics on interactive benchmarks.} We report dataset scale, semantic diversity, and scene complexity.}
\setlength{\tabcolsep}{8pt}
\label{tab:benchmark_stats}
\begin{tabular}{lcccccccc}
\toprule
\textbf{Benchmark} & \textbf{Scenes} &\textbf{ Room Cat.} & \textbf{Obj. Cat.} &\textbf{ Rooms/Sc.} & \textbf{Room Cat./Sc.} & \textbf{Obj./Sc. }& \textbf{Obj. Inst./Sc.} & \textbf{Contained Obj./Sc.} \\
\midrule
AI2-THOR   & 120  & 4   & 117 & 1.0 & 1.0 & 46.3 & 30.6 & 17.9 \\
OmniGibson & 45   & 39  & 237 & \textbf{7.1} &  \textbf{5.3} & 245.0 & 31.4 & 6.3 \\
InteriorGS & \textbf{1000} & \textbf{687} & \textbf{697} & 6.6 & 5.1 & \textbf{408.3} & \textbf{52.4} & \textbf{157.5} \textsuperscript{*} \\
\bottomrule
\end{tabular}
\begin{flushleft}
\footnotesize
\textsuperscript{*} Average computed only over scenes that have been annotated so far.
\end{flushleft}
\vspace{-0.2cm}
\end{table*}

A key design goal of \benchmark\ is to support multiple data sources without being tied to a specific simulator or dataset. By grounding the benchmark in 3D Scene Graph representations, we decouple the evaluation framework from any particular environment. To this end, we consider several existing interactive benchmarks and datasets that are directly compatible with our framework. In Tab.~\ref{tab:benchmark_stats}, we compare the diversity and complexity of three such compatible sources: AI2-THOR~\cite{kolve2017ai2}, OmniGibson~\cite{li2024behavior}, and InteriorGS~\cite{InteriorGS2025}. AI2-THOR consists exclusively of single-room scenes. OmniGibson includes larger scenes populated mostly with furniture and fixtures. Although additional objects can be sampled, doing so frequently results in failures and limits scalability. InteriorGS has a more diverse set of object categories, along with more fine-grained and descriptive room annotations. During manual annotation, we also ensure that realistic and diverse object–object relations are maintained. For these reasons, we use InteriorGS as the primary data source for \benchmark, while keeping the framework inherently data-agnostic and readily extensible to additional sources in the future.

\subsection{3D Scene-Graph Construction from Real Indoor Scans}
\label{sec:interiorgs}
\begin{figure*}[h]
\centering
\subfloat[]{
    \includegraphics[width=0.3\textwidth]{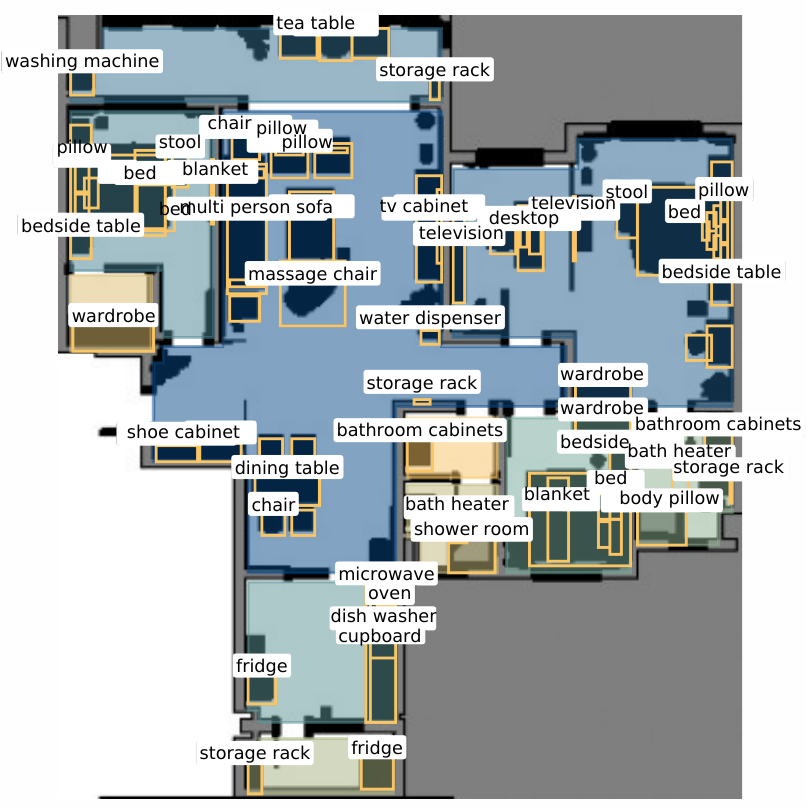}
}
\subfloat[]{
    \includegraphics[width=0.3\textwidth]{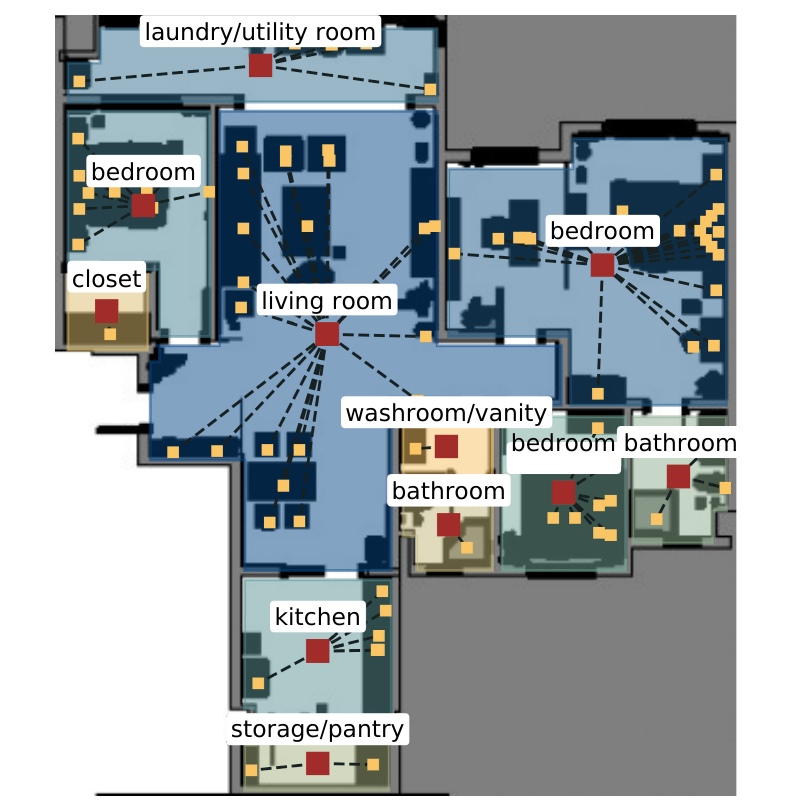}
}
\subfloat[]{
    \includegraphics[width=0.3\textwidth]{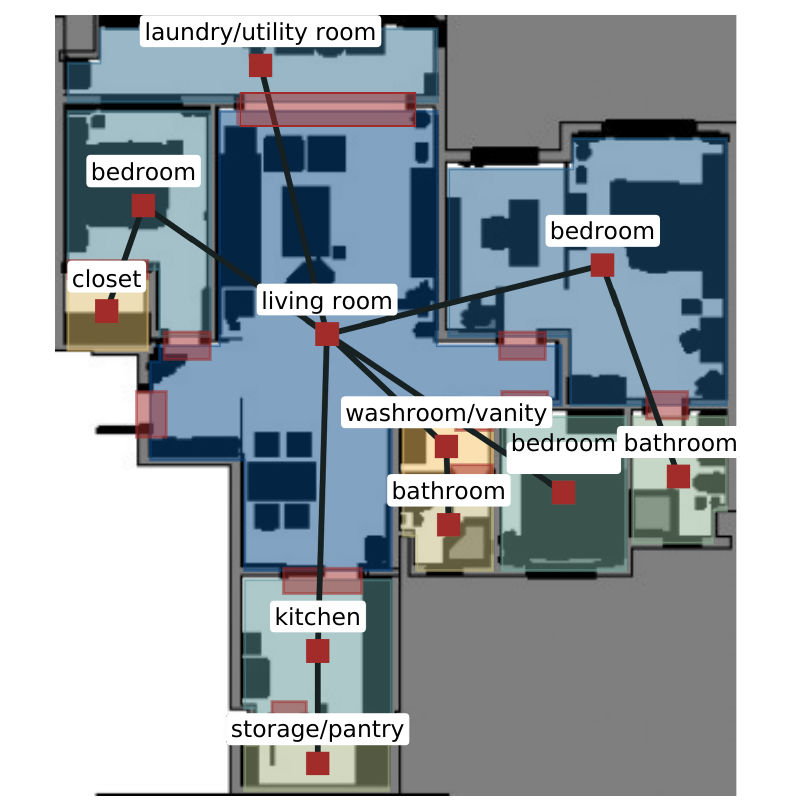}
}

\vspace{0.3em}

\subfloat[]{
    \includegraphics[width=0.3\textwidth]{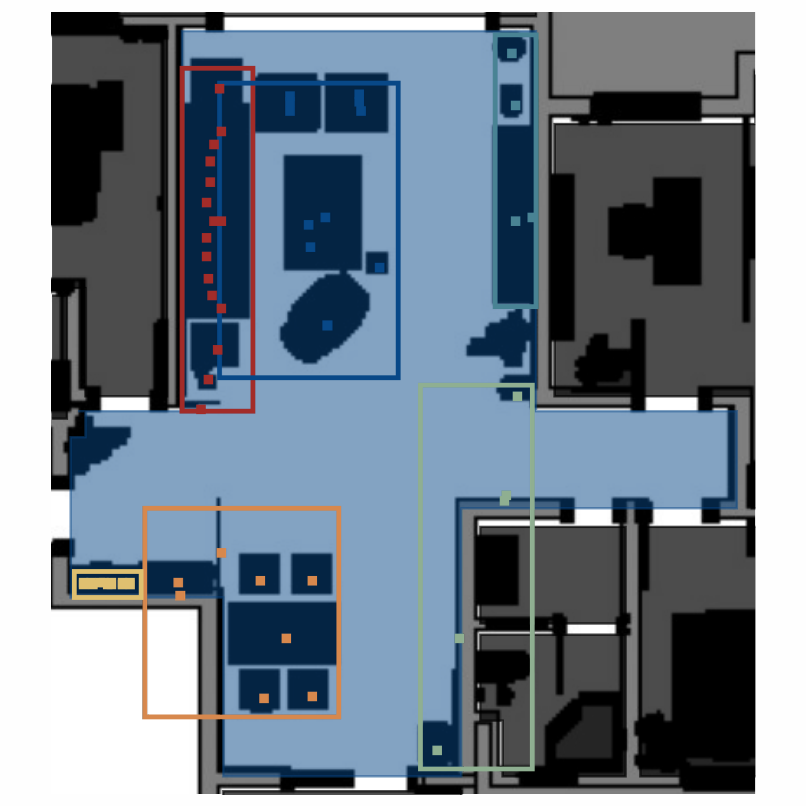}
}
\subfloat[]{
    \includegraphics[width=0.3\textwidth]{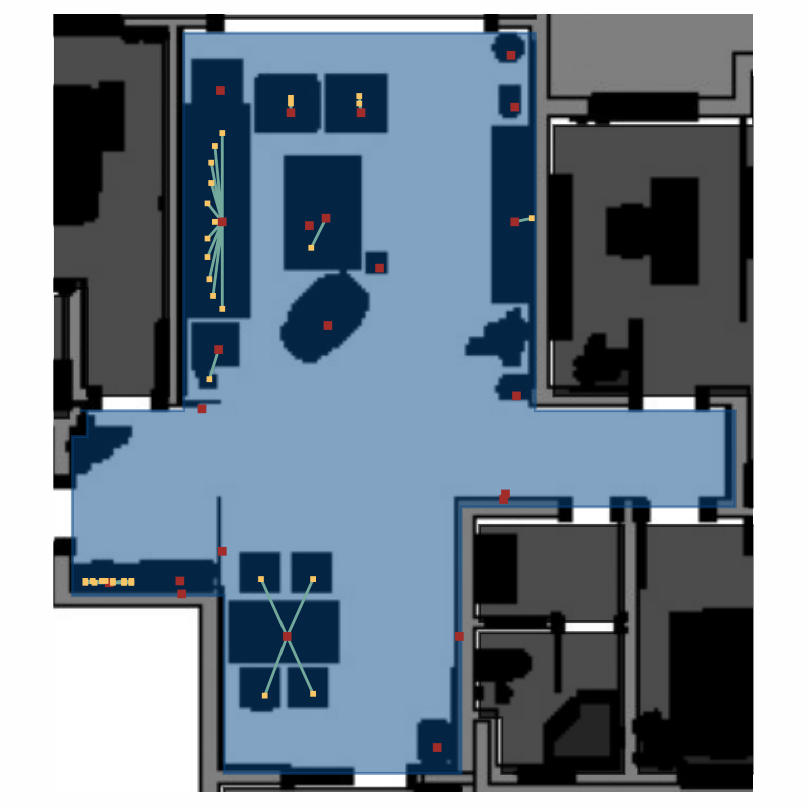}
}
\subfloat[]{
    \includegraphics[width=0.3\textwidth]{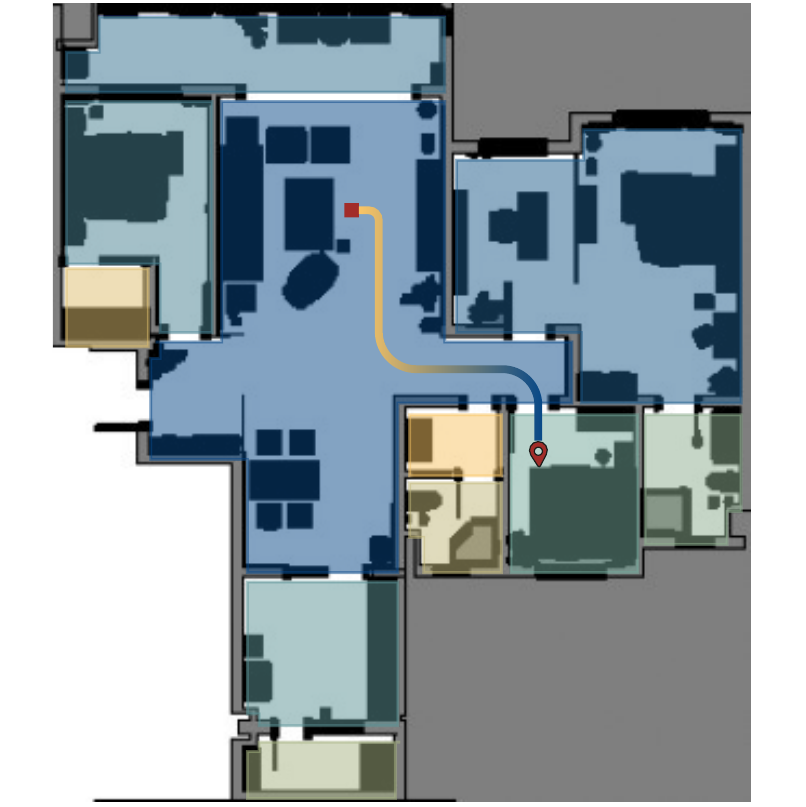}
}

    \caption{\textbf{3DSG construction pipeline from InteriorGS scenes.}  
    (a) Each scene includes a top-down occupancy map, annotated object bounding boxes, room polygons, and door/opening bounding boxes, which we use to construct the 3DSG.  
    (b) Each object's position is defined as the center of its bounding box. Objects are assigned to rooms if their centers lie within the corresponding room polygon. Rooms are annotated using an LLM based on their contents, and are then manually verified for consistency.  
    (c) To determine room connectivity, door polygons are dilated, and rooms sharing an intersecting door are marked as connected.
    (d) To define spatially coherent regions within rooms, we cluster object $(x, y)$ coordinates using a Gaussian Mixture Model. Each cluster represents a distinct region.  
    (e) Nested object relationships (e.g., \textit{on-top-of}, \textit{inside}) are manually annotated by experts by inspecting the Gaussian splats.  
    (f) Traveled distances are computed by first identifying the closest unoccupied pixel to the goal object within the same room on a dilated occupancy map, then applying A* to compute realistic geodesic paths from the agent to the goal.}
\label{fig:interiorgs}
\end{figure*}

In this environment, a scene is represented as a hierarchical 3DSG $\mathcal{G}$ described in Sec.~\ref{sec:problem}. To this end, we consider InteriorGS \cite{InteriorGS2025}, a Gaussian-splatting dataset of 1,000 human-centric indoor scenes (e.g., homes, restaurants, beauty salons) with dense object annotations. Besides its diversity, scene complexity, and large scale, this dataset provides an unbiased vocabulary distribution for evaluating our method. Since the raw dataset does not provide explicit 3DSG representations, we use a semi-automated extraction pipeline, illustrated in Fig.~\ref{fig:interiorgs},  to create them.

Each scene includes a Gaussian splat of the full environment, object bounding boxes with semantic labels, room and door polygons, and a top-down occupancy map. We extract 3DSG representations tailored to our environment as follows:\\
\textbf{Objects:} Each annotated object instance includes a semantic class and a 3D bounding box. We use the bounding box center as the node position and the semantic label as the object category.\\
\textbf{Rooms:} Room polygons are used to assign objects to rooms. Room classification is inferred via an LLM conditioned on room contents and manually refined to ensure accuracy. Door bounding boxes are used to derive room connectivity by detecting intersections between rooms and door polygons.\\
\textbf{Regions:} For each room, we cluster object positions $(x,y)$ using a Gaussian Mixture Model (GMM). The number of components is selected via the Bayesian Information Criterion (BIC) within predefined bounds. These clusters approximate spatially coherent regions that an agent would reveal during exploration. The region notation is used to mimic the agent's progressive discovery of nearby objects. It is purely organizational and does not affect the exploration methods evaluated.\\
\textbf{Nested Objects:} For object–object relations, we manually annotate \textit{on-top-of} and \textit{inside} edges using the underlying Gaussian splats, since bounding boxes alone produce unreliable relationships.

Furthermore, we use the occupancy map of the scene $M_{\mathcal{G}}$ to compute realistic traveled distances. To do so, we first sample the closest unoccupied point to the goal object within the same room and compute the geodesic path using Dijkstra’s algorithm \cite{dijkstra2022note} on the dilated occupancy map.

\subsection{Environment Roll-out}
The roll-out process, illustrated in Fig.~\ref{fig:rollout}, simulates how an agent incrementally explores a scene. Each episode begins by spawning the agent in a random region within a random room. The agent is provided with a query object $q$, an observed subgraph $\mathcal{G}_{\mathrm{obs}}$ consisting of objects and rooms previously observed, frontier nodes, i.e., regions whose children are not yet revealed, and unexplored room nodes, i.e., rooms connected via doors but not yet visited with all attributes hidden.

Actions correspond to selecting a node to explore as follows:
\begin{itemize}
    \item \textbf{Exploring an unexplored room:} reveals the closest region relative to the agent, all first-level objects in that region, and any additional connected frontiers or rooms.
    \item \textbf{Exploring a frontier:} reveals the first-level objects in the corresponding region.
    \item \textbf{Exploring an object:} reveals its nested objects (e.g., objects inside containers or supported by the object).
\end{itemize}

The episode terminates when the query object $q$ becomes visible to the agent or when a maximum number of steps $N_{\max}$ is reached. An episode is successful if the agent reveals the goal object within $N_{\max}$ steps.

\section{Experimental Evaluation}
Our experiments are designed to answer the following:
\begin{itemize}
\item Do our learned utility scoring functions improve search efficiency compared to embedding similarity?
\item Can our method match the performance of online LLM-based planners with a fraction of the runtime?
\item How do different components of our approach influence performance, and which choices matter most?
\item Does our method transfer to real-world applications?
\item How robust is our method under real-world conditions, including the presence of real-time perception noise and segmentation errors?
\end{itemize}
\label{sec:eval}

\subsection{Implementation Details}
\label{sec:implementation}
\begin{figure*}
    \centering
    \includegraphics[width=\linewidth]{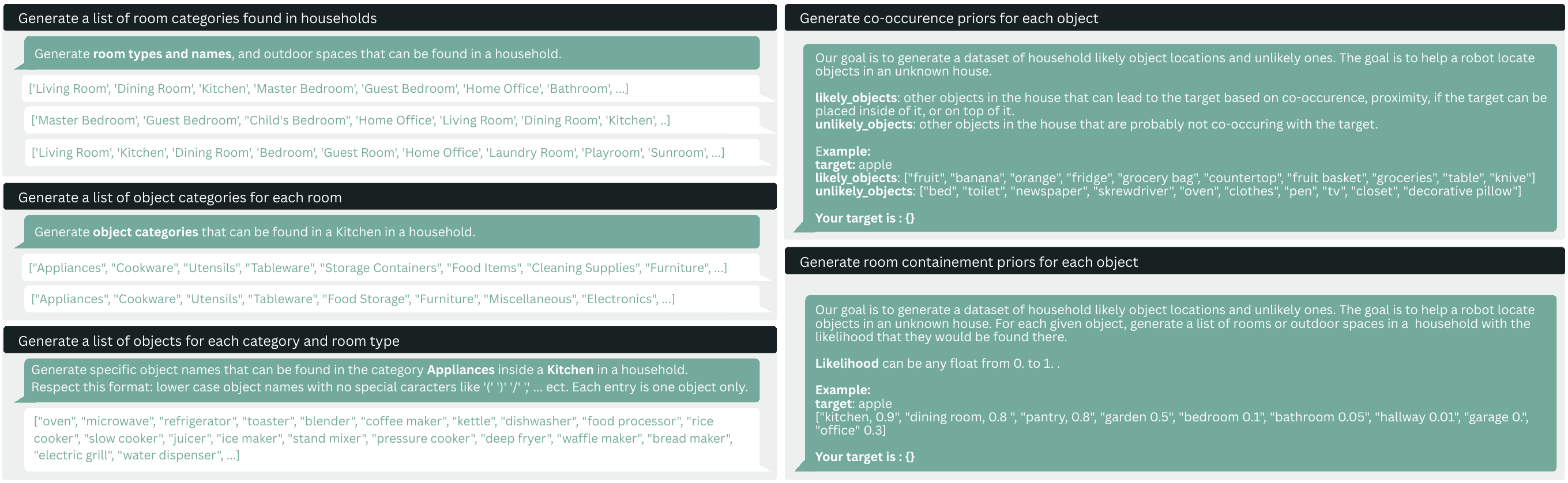}
    \caption{\textbf{Prompts used for procedural relational semantic knowledge distillation.}
    \textbf{Left:} Prompts used to generate a large and diverse set of household objects; (top) household room categories, (middle) object categories per room, and (bottom) object set associated with each room–object category pair.
    \textbf{Right:} Prompts used to generate relational semantic datasets for object–object co-occurrence and room–object containment.}
    \label{fig:all_prompts}
\end{figure*}

As described in Sec.~\ref{sec:knowledge_distillation}, we procedurally generate two supervised learning datasets using the \texttt{gpt-4o-mini} model. This model offers a trade-off between cost and performance and supports structured outputs, thereby simplifying parsing and reducing errors caused by hallucinations or format violations. The generated data span $N_{\mathrm{objects}}=\texttt{3824}$ unique object categories, with dataset sizes of \texttt{92K} samples for containment and \texttt{7.5K} samples for co-occurrence. Dataset generation required approximately three hours per dataset. We provide examples of the prompts used in Fig.~\ref{fig:all_prompts}.

We then use a frozen SBERT embedder (22.7M parameters) to vectorize the inputs as described in Eq.~\ref{eq:vectorize}, resulting in vectors of size $384\times2=768$ and train two 3-layer MLPs, with layers (512, 256, 128) each, totaling 558K parameters per relational model. For all symbolic, simulation, and real-world experiments, we used the same model checkpoints trained as described in Sec.~\ref{sec:implementation}. Unless stated otherwise, the agent's hyperparameters are fixed across all experiments as shown in Tab.~\ref{tab:hyp}.

 \begin{table}[h]
    \centering
    \caption{\textbf{Hyperparameters used to instantiate our agent.}}
    \begin{tabular}{l c}
        \toprule
        \textbf{Hyperparameter} & \textbf{Value} \\ 
        \midrule 
        Default room exploration score & 0.7 \\ 
        Default frontier exploration score & 0.6 \\ 
        $\Delta$ (utility margin) & 0.1 \\ 
        Room influence weight $w$ & 0.3 \\ 
        \bottomrule 
    \end{tabular}
    \label{tab:hyp}
\end{table}

\subsection{Limitations of Embedding Similarity} 
\label{sec:limitations}
\begin{figure}[t]
    \centering
    \includegraphics[width=\linewidth]{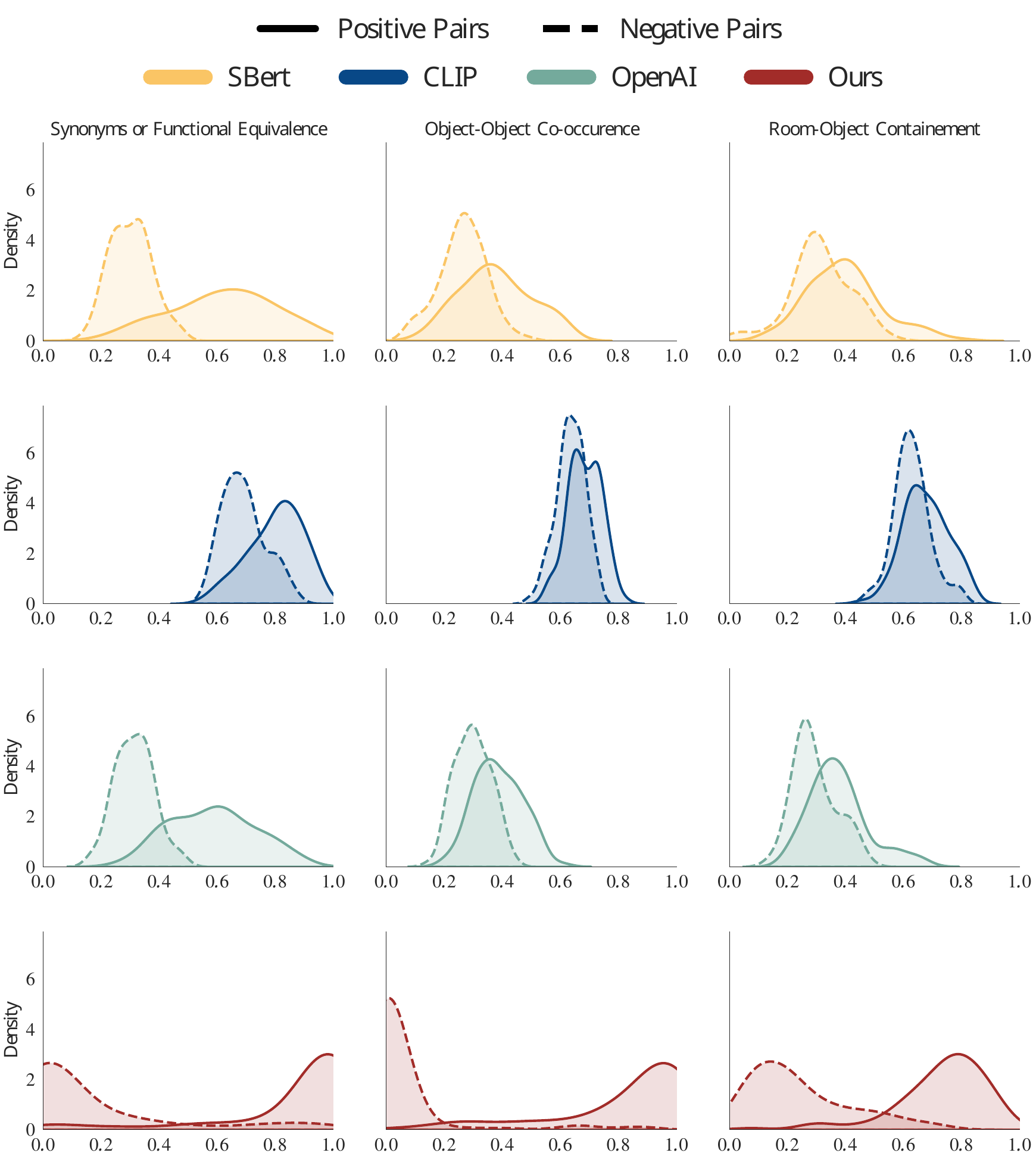}
    \caption{\textbf{Comparison of embedding similarity distributions with our learned relational scoring models.} Synonym pairs are separable across all models, whereas co-occurrence and containment relationships are not. Our learned models produce substantially stronger separation.}
    \label{fig:limitations}
\end{figure}

Embeddings are widely used to compute similarity between a query and observations (e.g., a language query and RGB images), and the resulting similarity scores are then used to guide downstream tasks such as object search. While effective in measuring visual or conceptual similarity, they fall short in encoding relational semantic information critical for reasoning and decision-making. To illustrate these limitations, we analyze the similarity distributions of different relational semantics using pretrained text-only embedders including SBert~\cite{reimers2019sentence} and OpenAI vector embeddings as well as a vision–language embedder CLIP~\cite{radford2021learning}, and compare them with our learned scoring functions $f_{\theta_1}^{co\text{-}occur}$ and $f_{\theta_2}^{contain}$. We consider three categories: (a) synonyms or functionally equivalent objects, (b) object–object co-occurrence, and (c) room–object containment. We use $f_{\theta_1}^{co\text{-}occur}$ for the first two categories, as functionally equivalent items also tend to co-occur. For each category, we manually curated 100 positive pairs and 100 negative pairs. Tab.~\ref{tab:pos_and_neg_pairs} shows representative examples of the pairs used to compare the resulting similarity scores.
Fig.~\ref{fig:limitations} shows the similarity and our learned scores distributions across these categories. While synonyms (a) are generally well separated, the distributions for co-occurrence (b) and room–object containment (c) show significant overlap between positive and negative pairs. This indicates that current embedding similarity-based methods lack the discriminative power to encode relational semantics. In contrast, our learned relevance models, which are explicitly trained on co-occurrence and containment relations, produce clearly separated distributions.

\begin{table*}
\centering
\setlength{\tabcolsep}{20pt}
\caption{\textbf{Examples of positive and negative pairs used to evaluate relational semantics.}}
\label{tab:pos_and_neg_pairs}

\begin{tabular}{l ll ll} 
\toprule 
\textbf{Relational Semantic} & \multicolumn{2}{c}{\textbf{Positive Pairs}} & \multicolumn{2}{c}{\textbf{Negative Pairs}} \\ 
\midrule 
\multirow{2}{*}{Synonyms and Functionally Equivalent} & \cellcolor{gray!15}sponge & \cellcolor{gray!15}scrubber & \cellcolor{gray!15}tv & \cellcolor{gray!15}plant \\ 
& laptop & notebook & bottle & car \\ 
& \cellcolor{gray!15}bookshelf & \cellcolor{gray!15}bookcase & \cellcolor{gray!15}broom & \cellcolor{gray!15}clock \\ 
& game console & video game system & fridge & chair \\ 
& \cellcolor{gray!15}ironing board & \cellcolor{gray!15}iron stand & \cellcolor{gray!15}fan & \cellcolor{gray!15}watch \\ 
\midrule 
\multirow{3}{*}{Co-occur} & shower & towel & kettle & bed \\ 
& \cellcolor{gray!15}router & \cellcolor{gray!15}desk & \cellcolor{gray!15}book & \cellcolor{gray!15}toilet \\ 
& rug & couch & pillow & desk \\ & \cellcolor{gray!15}can opener & \cellcolor{gray!15}drawer & \cellcolor{gray!15}nightstand & \cellcolor{gray!15}microwave \\ 
& coffee table & couch & hanger & dishwasher \\ 
\midrule 
\multirow{3}{*}{Contain} & \cellcolor{gray!15}magazine & \cellcolor{gray!15}living room & \cellcolor{gray!15}whiteboard & \cellcolor{gray!15}bathroom \\ 
& frame & living room & toothbrush & dining room \\ 
& \cellcolor{gray!15}oven & \cellcolor{gray!15}kitchen & \cellcolor{gray!15}dresser & \cellcolor{gray!15}bathroom \\ 
& drawer & bedroom & bathtub & laundry room \\ 
& \cellcolor{gray!15}toilet & \cellcolor{gray!15}bathroom & \cellcolor{gray!15}sofa & \cellcolor{gray!15}kitchen \\ 
\bottomrule 
\end{tabular} 
\end{table*}

\subsection{Benchmarks}
In this section, we provide more details and examples of the evaluation benchmarks we used in Sec.~\ref{sec:eval}.
We evaluate our approach on two environments:

\noindent\textbf{Symbolic Benchmark}: We evaluate in our \benchmark benchmark over 200 episodes across 10 indoor household scenes from the InteriorGS dataset. Each episode specifies a query object located within one of the annotated scenes. To simulate interactive search, we assume that all nested objects are either inside or on top of another object, requiring the agent to explore the parent object to locate them. For non-interactive scenarios, we assume the target object is visible when exploring the region of the room where it is located. We use 142 unique open-vocabulary object queries, including 61 categories that do not exist in the training data used in Sec.~\ref{sec:implementation}. Among these, 115 target objects are not immediately observable and must be discovered through interaction with other objects. We set the maximum episode length to $N_{\max}=50$ steps.  Tab.~\ref{tab:episodes_symsearch} provides descriptions of evaluation episodes used in our experiments.

\noindent\textbf{Simulation Benchmark}: We extend the OmniGibson simulator with an interactive object search task definition. Target objects are spawned using the simulator's internal object sampler, either inside containers or within rooms. Each episode is defined by the query object category, an optional parent object and predicate, and the room in which the object should be spawned. If a parent object is specified, we first check whether it exists in the target room. If it does not exist, we sample the parent object in the scene before sampling the target object according to the specified predicate. If no parent object is specified, we first check whether the target object exists in the room. If it does, we use it as the target. Otherwise, we sample it in that room. If no room is specified, we select any instance of the target category in the scene or randomly sample one if none exists.
We generate 50 episodes, each containing 50 unique BEHAVIOR-1K objects across 11 scenes. Among these, 27 objects are not included in the training data, and 25 require interactivity. To fairly evaluate semantic reasoning performance between baselines, the agent uses an oracle “magic open” manipulation action and ground truth segmentation to avoid failure cases caused by manipulation or perception. We set $N_{\max}=35$ steps. 

Success is defined as the agent having the target object visible and within a distance threshold. As metrics, we compute the Success Rate (SR) and Success weighted by Path Length (SPL). Additionally, we record the number of high-level steps taken per episode and the average inference time to quantify exploration efficiency.  Tab.~\ref{tab:episodes_omnigibson} provides descriptions of evaluation episodes used in our experiments.

\begingroup
\rowcolors{2}{gray!15}{white}

\begin{table*}[ht]
\centering
\caption{\textbf{Examples of evaluation episodes from \benchmark}}
\label{tab:episodes_symsearch}

\setlength{\tabcolsep}{6pt}

\begin{tabular}{p{3.2cm} p{3.2cm} >{\centering\arraybackslash}p{2cm} p{7.0cm}}
\hline
\textbf{Query} & \textbf{Scene} & \textbf{Interactive} & \textbf{Location Description} \\
\hline

teapot & 0900 & \checkmark & on top of a teatable in a combined living room and kitchen \\
refrigerator & 0900 & $\times$ & in a combined living room and kitchen \\
vegetable & 0901 & \checkmark & on top of a chopping board in a kitchen \\
eggbeater & 0902 & \checkmark & inside a cabinet in a kitchen \\
tv & 0902 & $\times$ & in a bedroom \\
multi person sofa & 0902 & $\times$ & in a living room \\
keyboard & 0904 & \checkmark & on top of a table in a bedroom \\
chopsticks & 0906 & \checkmark & inside a storage rack in a kitchen \\
faucet & 0906 & \checkmark & on top of a washing station in a bathroom \\
table lamp & 0906 & $\times$ & in a combined living and dining room \\
headphones & 0908 & \checkmark & inside a display cabinet in a combined study and library \\
rice cooker & 0909 & \checkmark & inside a storage rack in a kitchen \\
\hline

\end{tabular}
\end{table*}
\endgroup

\begingroup
\rowcolors{2}{gray!15}{white}

\begin{table*}[ht]
\centering
\caption{\textbf{Examples of evaluation episodes from OmniGibson}}
\label{tab:episodes_omnigibson}

\setlength{\tabcolsep}{6pt}

\begin{tabular}{p{3.2cm} p{3.2cm} >{\centering\arraybackslash}p{2cm} p{7.0cm}}
\hline
\textbf{Query} & \textbf{Scene} & \textbf{Interactive} & \textbf{Location Description} \\
\hline

chopstick & Pomaria\_1\_int & $\times$ & on top of a breakfast table in a kitchen \\
wall\_mounted\_tv & Wainscott\_0\_int & $\times$ & on top of a console table in a living room \\
lipstick & Beechwood\_0\_int & $\times$ & on top of a sink in a bathroom \\
vacuum & Beechwood\_0\_int & $\times$ & in a utility room \\
keyboard & Beechwood\_0\_int & $\times$ & on top of a desk in a private office \\
dental\_floss & Merom\_0\_int & \checkmark & inside of a bottom cabinet in a bathroom \\
pants & Beechwood\_1\_int & \checkmark & inside of a hamper in a bedroom \\
box\_of\_sanitary\_napkins & Beechwood\_1\_int & \checkmark & inside of a top cabinet in a bathroom \\
measuring\_cup & Merom\_1\_int & \checkmark & inside of a bottom cabinet in a kitchen \\
\hline

\end{tabular}
\end{table*}
\endgroup

\subsection{Baselines}
We compare our agent against the following baselines:
\paragraph{Random agent}: The agent uniformly samples actionable nodes.
\paragraph{Embedding-based agents}: Current baselines that use embedding similarity to guide the search do not support interactive exploration. They either select frontiers based on similarity scores or learn a navigation policy conditioned on embeddings. To allow for a fair comparison, we instantiate these agents identically to ours, but replace the learned approximation priors in Eq.~\ref{eq:p_contain} and Eq.~\ref{eq:p_cooccur} with a similarity score between the query and scene graph embeddings using  SBERT~\cite{reimers2019sentence} or CLIP~\cite{radford2021learning}. Formally, the utility of a node is computed as:
    \begin{equation}
    \label{eq:sbert_clip_room}
    u_q(r) \approx \mathrm{sim}(E(q), E(r)),
    \end{equation}
    for rooms, and
    \begin{equation}
    \label{eq:sbert_clip_object}
    u_q(o) \approx \mathrm{sim}(E(q), E(o))
    \end{equation}
    for objects, where $E(\cdot)$ denotes the node embedding (from SBERT or CLIP), and $\mathrm{sim}(\cdot,\cdot)$ is cosine similarity.
    Since the similarity distributions vary across models (see Figure~\ref{fig:limitations}), we manually tune the default room and frontier exploration scores to account for the distributional shift as summarized below:
    \begin{table} [H]
        \centering
        \caption{\textbf{Hyperparameters used for Embeddings-based agents.}}
        \begin{tabular}{l l c} 
            \toprule 
            \textbf{Embedder} & \textbf{Hyperparameter} & \textbf{Value} \\ 
            \hline 
            \multirow{4}{*}{SBERT} 
                & Default room exploration score & 0.3 \\ 
                & Default frontier exploration score & 0.25 \\
                & $\Delta$ (utility margin) & 0.05 \\ 
                & Room influence weight $w$ & 0.3 \\ 
            \hline
            \multirow{4}{*}{CLIP} 
                & Default room exploration score & 0.7 \\ 
                & Default frontier exploration score & 0.4 \\ 
                & $\Delta$ (utility margin) & 0.05 \\ 
                & Room influence weight $w$ & 0.3 \\ 
            \bottomrule
        \end{tabular}
\end{table}
\paragraph{LLM-Based Agents}
We compare our method against two approaches that use online LLM planners for interactive object search. Both methods share the same pipeline as our agent—namely, constructing a 3D scene graph (3DSG) from raw observations and mapping high-level actions to low-level policies, and differ only in how the next node to explore is selected. We describe each approach below.
\begin{itemize}
    \item \textbf{MoMa-LLM}~\cite{honerkamp2024language}: The agent queries an LLM at every high-level decision step to select the next action.  We follow the prompt format proposed in the original work~\cite{honerkamp2024language}. To reduce hallucinations, we use structured outputs that constrain the model to select only valid actions and arguments. The LLM is also provided with a history of the previous 30 actions and their outcomes to discourage repeated selection of the same actions.
    \item \textbf{GODHS}~\cite{zhang2025language}: This method uses a different approach that significantly reduces the number of LLM calls performed during online exploration.  The agent first explores all rooms and frontiers in the environment. It then queries the LLM to rank the visited rooms by their relevance to the target object and visits them in that order. Within each room, the LLM is first asked to identify potential carriers for the target object and then to rank these carriers by the likelihood of containing the queried object. While this approach improves runtime efficiency by limiting repeated LLM calls, it introduces an additional exploration phase.
\end{itemize}

To select the LLM variant for these baselines, we conducted comparative evaluations across various LLMs, including open-source alternatives. We provide the full details and results in Sec.~\ref{sec:ablations}. Consistent with existing literature~\cite{rana2023sayplan,honerkamp2024language,mohammadi2025more}, we found that the later OpenAI GPT series~\cite{achiam2023gpt} offer the best compromise between inference time and performance on reasoning and planning tasks. We opted for the "gpt-5 mini" model for both agents due to its ability to handle structured output and offer a compromise between performance and cost.

\subsection{Evaluation Results}

\begin{table*}[ht]
\centering
\caption{\textbf{Evaluation results on \benchmark and Omnigibson.} Table reports the Success Rate (SR), Success weighted by Path Length (SPL), average number of steps (high-level actions) per episode, and average inference time in seconds. Results on SymSearch are averaged across 5 different seeds. We report the \besttext{best}, \secondtext{second-best}, and \thirdtext{third-best} results for each metric, respectively.}
\setlength{\tabcolsep}{4pt}
\footnotesize
\begin{tabular}{llcccccccc}
\toprule
& & \multicolumn{4}{c}{\textbf{SymSearch}} & \multicolumn{4}{c}{\textbf{OmniGibson}} \\
\cmidrule(lr){3-6} \cmidrule(lr){7-10}
\textbf{Method} & \textbf{Agent} & SR $\uparrow$ & SPL $\uparrow$ & N Steps $\downarrow$ & Inference & SR $\uparrow$ & SPL $\uparrow$ & N Steps $\downarrow$ & Inference \\
&  &  &  &  & Time~(s)$\downarrow$ &  &  &  & Time~(s)$\downarrow$ \\
\midrule
\multirow{1}{*}{Random} 
& Random Agent & $0.381 \pm 0.043$ & $0.074 \pm 0.012$ & $45 \pm 2$ & $0.00 \pm 0.00$ & - & - & - & - \\
\midrule
\multirow{2}{1cm}{Embedding Similarity} 
& CLIP Similarity (w/o RI) & $0.564 \pm 0.010$ & $0.172 \pm 0.018$ & $37 \pm 1$ & $0.58 \pm 0.01$ & - & - & - & - \\
& CLIP Similarity & $0.650 \pm 0.006$ & $0.177 \pm 0.019$ & $33 \pm 1$ &  $0.75 \pm 0.01$ & $0.565$ & $0.322$ & $23$ & \secondcell $1.5$ \\
& SBERT Similarity (w/o RI) & $0.632 \pm 0.029$ & $0.180 \pm 0.017$ & $36 \pm 1$ & \bestcell $0.08 \pm 0.00$ & - & - & - & - \\
& SBERT Similarity & $0.689 \pm 0.007$ & $0.192 \pm 0.014$ & $32 \pm 0$ & \thirdcell $1.00 \pm 0.01$ & $0.632$ & $0.241$ & $24$ & \bestcell $0.1$ \\
\midrule
\multirow{2}{*}{LLM-based} 
& MoMa-LLM~\cite{honerkamp2024language} & $0.827 \pm 0.012$ & \thirdcell $0.255 \pm 0.020$ & \secondcell $21 \pm 1$ & $294.52 \pm 20.57$ & $0.696$ & $0.257$ & \bestcell $17$ & $300$ \\
& GODHS~\cite{zhang2025language} & \bestcell $0.906 \pm 0.011$ & $0.159 \pm 0.016$ & \bestcell $20 \pm 0$ & $39.28 \pm 2.17$ & $0.478$ & $0.087$ & $26$ & \thirdcell $35$ \\
\midrule
\multirow{4}{*}{Ours} 
& \methodname (w/o RI, $\Delta=0$) & $0.834 \pm 0.004$ & $0.215 \pm 0.013$ & \thirdcell $22 \pm 0$ & \bestcell $0.08 \pm 0.00$ &  $0.745$ & \thirdcell $0.340$ & \thirdcell $23$ & $0.1$ \\
& \methodname (w/o RI) & $0.840 \pm 0.006$ & \secondcell $0.264 \pm 0.023$ & $23 \pm 1$ & \secondcell $0.09 \pm 0.00$ & \thirdcell $0.756$ & $0.336$ & $23$ & \bestcell $0.1$ \\
& \methodname ($\Delta=0$) & \thirdcell  $0.871 \pm 0.002$ & $0.219 \pm 0.012$ & \bestcell $20 \pm 0$ & \secondcell $0.09 \pm 0.00$ & \secondcell $0.826$ & \secondcell $0.413$ & \secondcell $18$ & \bestcell $0.1$ \\
& \methodname & \secondcell $0.876 \pm 0.002$ & \bestcell $0.282 \pm 0.021$ & \secondcell $20 \pm 1$ & \bestcell $0.08 \pm 0.00$ & \bestcell $0.829$ & \bestcell $0.415$ & \thirdcell $19$ & \bestcell $0.1$ \\
\bottomrule
\multicolumn{10}{l}{\scriptsize RI = Room Influence.} \\
\end{tabular}
\label{tab:evaluation_results}
\end{table*}

Tab.~\ref{tab:evaluation_results} summarizes results on both benchmarks. On the symbolic benchmark, embedding-based baselines outperform the random agent but remain well below stronger methods. GODHS achieves a high SR but a low SPL due to an exhaustive exploration phase. This means that the agent first gathers more information about the environment by identifying all possible rooms in the household, which leads to longer travel distances but enables more informed decisions at later stages.
 MoMa-LLM achieves competitive SPL but relies on costly LLM calls at every step, resulting in inference costs that are four orders of magnitude higher. Our method consistently outperforms embedding-based agents in both SR and SPL, while matching LLM-based SR at a fraction of the computational cost. Ablations show that room influence has the largest impact on SR, while removing the utility margin reduces efficiency, as the agent greedily selects high-scoring nodes without accounting for traveled distance. We also compare two variants of the embedding-based baselines, with and without room influence, to highlight the impact of this component on performance. We show that adding room context improves the overall success rate, even when utility scores are noisy.

\begin{figure}
    \centering
    \includegraphics[width=\linewidth]{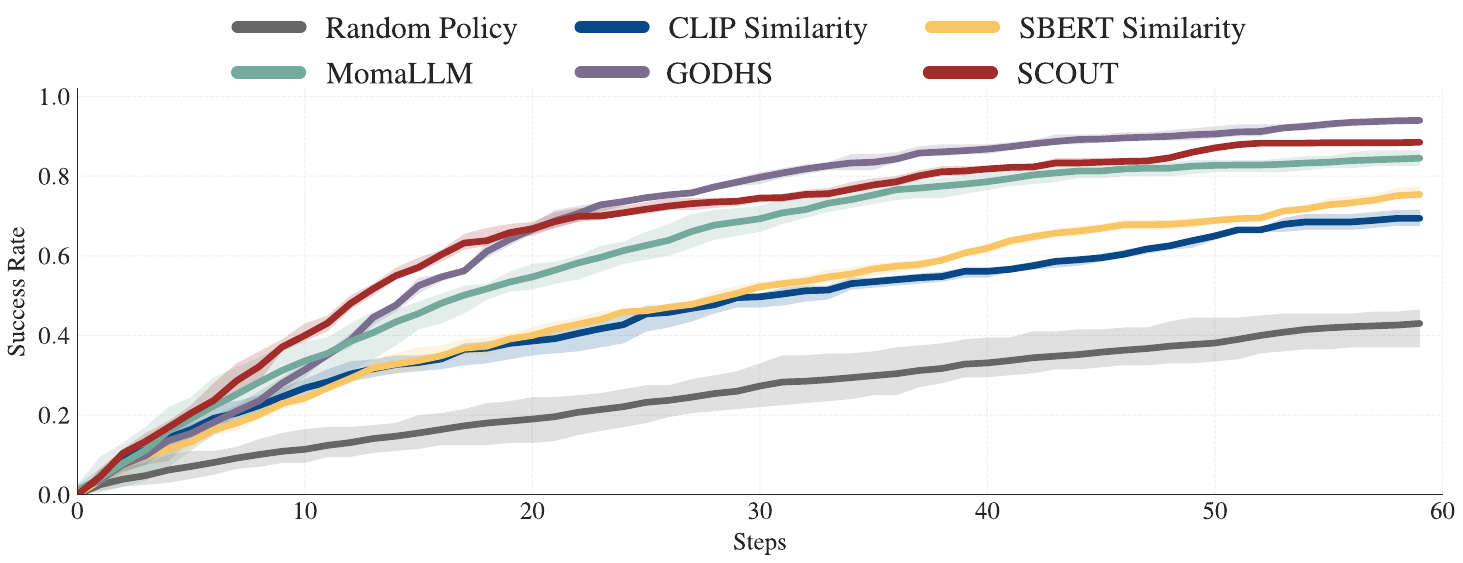}
    \caption{\textbf{Success rate over an increasing number of steps on \benchmark.} Our method closely follows the performance of LLM-based baselines throughout the roll-out.}
    \label{fig:auc_symsearch}
\end{figure}

Fig.~\ref{fig:auc_symsearch} shows the success rate over an increasing number of steps. Throughout the roll-out, our agent closely matches and often surpasses LLM-based planners, particularly in early stages, where it effectively balances exploration and exploitation. In contrast, GODHS initially over-explores rooms, while MoMa-LLM exhibits higher variance and slightly lower SR due to non-determinism and hallucinations. We hypothesize that LLMs implicitly leverage similar priors in decision-making, which we attribute to our agent’s ability to mimic their behavior.

Contrary to \benchmark where agents have privileged information of when all rooms have been explored, the simulation benchmark poses more challenges that can arise from the real-time scene exploration and scene graph construction. This means that an efficient strategy must balance exploration and exploitation. As a result, GODHS performance drops as it over-explores all frontiers and closed doors. MoMa-LLM performs slightly better but also tends to prefer exploring unexplored space over attempting to open containers. 
In both evaluations, our agent balanced exploration and exploitation for maximum efficiency, surpassing the performance of LLM-based planners while remaining up to orders of magnitude faster at inference.

\subsection{Qualitative Analysis of the Learned Utility Scores}
To illustrate how the previously distilled relational semantic models guide exploration, we report example utility scores computed for a set of rooms and objects, along with their updated scores defined in Eq.~\ref{eq:updated_scores}. We consider multiple queries, including both in-distribution queries (highlighted in green) and out-of-distribution queries (highlighted in yellow and blue). Tab.~\ref{tab:example_scores} summarizes these examples. We observe that the influence of rooms on object scores is crucial, as it allows the model to semantically differentiate objects and their uses based on their locations. For instance, cabinets and drawers are commonly found in different rooms, each typically containing specific target objects, which naturally gives them high scores. Without accounting for room context, a cabinet in the kitchen and a cabinet in the bathroom would receive the same score for a query like “plate.” By incorporating the room score, we add semantic context, reducing the score of a container if its associated room is not relevant to the target. Interestingly, because the relational models receive pretrained text embeddings as input, they generalize to descriptive queries (highlighted in blue) such as `something to cook with' or `something to sit on', even though no similar queries are seen during training.
\begin{table}
\centering
\footnotesize
\caption{\textbf{Example utility scores for room and objects categories across multiple queries.}
In-distribution queries are highlighted in \textcolor{bestgreen}{green}, out-of-distribution queries in \textcolor{secondyellow}{yellow}, and descriptive queries in \textcolor{thirdblue}{blue}.}
\label{tab:example_scores}

\begin{tabular}{ c l c l c c }

\toprule
\textbf{Query} & \makecell{\textbf{Room} \\ \textbf{Type}} & \makecell{\textbf{Room} \\ \textbf{Score}} &
\textbf{Object} & \makecell{\textbf{Object} \\ \textbf{Score}} & \makecell{\textbf{Updated} \\ \textbf{Score}} \\
\midrule

\multirow{17}{*}{\textcolor{bestgreen}{plate}} 
& \multirow{6}{*}{kitchen} & \multirow{6}{*}{0.78} & counter & 0.99 & 0.78 \\
&  &  & cabinet & 0.94 & 0.75 \\
&  &  & sink & 0.94 & 0.75 \\
&  &  & table & 0.92 & 0.74 \\
&  &  & fridge & 0.45 & 0.48 \\
&  &  & chair & 0.12 & 0.30 \\
\cline{2-6}
& \multirow{4}{*}{living room} & \multirow{4}{*}{0.30} & cabinet & 0.94 & 0.29 \\
&  &  & coffee table & 0.82 & 0.26 \\
&  &  & tv & 0.04 & 0.10 \\
&  &  &  sofa & 0.00 & 0.09\\
\cline{2-6}
& \multirow{3}{*}{bathroom} & \multirow{3}{*}{0.21} & sink & 0.94 & 0.20 \\
&  &  & cabinet & 0.94 & 0.20 \\
&  &  & bathtub & 0.00 & 0.06 \\
\hline

\multirow{12}{*}{\textcolor{bestgreen}{toothpaste}} 
& \multirow{3}{*}{bathroom} & \multirow{3}{*}{0.88} & cabinet & 1.00 & 0.88 \\
&  &  & sink & 0.99 & 0.88 \\
&  &  & bathtub & 0.19 & 0.39 \\
\cline{2-6}
& \multirow{4}{*}{bedroom} & \multirow{4}{*}{0.55} & cabinet & 1.00 & 0.55 \\
&  &  & nightstand & 0.99 & 0.55 \\
&  &  & bed & 0.01 & 0.17 \\
\cline{2-6}
& \multirow{5}{*}{kitchen} & \multirow{5}{*}{0.36} & cabinet & 1.00 & 0.36 \\
&  &  & sink & 0.99 & 0.36 \\
&  &  & table & 0.88 & 0.33 \\
&  &  & fridge & 0.10 & 0.13 \\
\hline

\multirow{9}{*}{\textcolor{secondyellow}{lip gloss}} 
& \multirow{3}{*}{bathroom} & \multirow{3}{*}{0.65} & cabinet & 0.99 & 0.65 \\
&  &  & sink & 0.83 & 0.58 \\
&  &  & bathtub & 0.00 & 0.20 \\
\cline{2-6}
& \multirow{4}{*}{bedroom} & \multirow{4}{*}{0.58} & nightstand & 0.99 & 0.58 \\
&  &  & cabinet & 0.99 & 0.58 \\
&  &  & wardrobe & 0.28 & 0.29 \\
\cline{2-6}
& \multirow{3}{*}{kitchen} & \multirow{3}{*}{0.37} & counter & 1.00 & 0.37 \\
&  &  & cabinet & 0.99 & 0.37 \\
&  &  & fridge & 0.03 & 0.12 \\
\hline

\multirow{9}{*}{\textcolor{secondyellow}{melon}} 
& \multirow{5}{*}{kitchen} & \multirow{5}{*}{0.97} & counter & 0.99 & 0.97 \\
&  &  & cabinet & 0.92 & 0.92 \\
&  &  & table & 0.73 & 0.79 \\
&  &  & fridge & 0.33 & 0.51 \\
&  &  & sink & 0.09 & 0.35 \\
\cline{2-6}
& \multirow{2}{*}{living room} & \multirow{2}{*}{0.42} & coffee table & 0.73 & 0.34 \\
&  &  & sofa & 0.00 & 0.13 \\
\cline{2-6}
& \multirow{2}{*}{bathroom} & \multirow{2}{*}{0.17} & cabinet & 0.92 & 0.16 \\
&  &  & sink & 0.09 & 0.06 \\
\hline

\multirow{10}{*}{\textcolor{thirdblue}{\makecell{a brown \\ leather \\ jacket}}} 
& \multirow{4}{*}{bedroom} & \multirow{4}{*}{0.80} & wardrobe & 0.99 & 0.79 \\
&  &  & nightstand & 0.92 & 0.75 \\
&  &  & cabinet & 0.87 & 0.72 \\
&  &  & bed & 0.07 & 0.28 \\
\cline{2-6}
& \multirow{4}{*}{living room} & \multirow{4}{*}{0.72} & cabinet & 0.87 & 0.66 \\
&  &  & coffee table & 0.42 & 0.43 \\
&  &  & tv & 0.01 & 0.22 \\
&  &  & sofa & 0.01 & 0.22 \\
\cline{2-6}
& \multirow{2}{*}{bathroom} & \multirow{2}{*}{0.10} & cabinet & 0.87 & 0.09 \\
&  &  & bathtub & 0.00 & 0.03 \\
\hline

\multirow{12}{*}{\textcolor{thirdblue}{\makecell{a snack \\ to eat}}}  
& \multirow{5}{*}{kitchen} & \multirow{5}{*}{0.62} & counter & 1.00 & 0.62 \\
&  &  & table & 0.96 & 0.60 \\
&  &  & cabinet & 0.96 & 0.60 \\
&  &  & fridge & 0.80 & 0.53 \\
&  &  & sink & 0.01 & 0.19 \\
\cline{2-6}
& \multirow{4}{*}{living room} & \multirow{4}{*}{0.53} & coffee table & 0.98 & 0.52 \\
&  &  & sofa & 0.00 & 0.16 \\
&  &  & tv & 0.00 & 0.16 \\
&  &  & cabinet & 0.96 & 0.51 \\
\cline{2-6}
& \multirow{3}{*}{bathroom} & \multirow{3}{*}{0.09} & cabinet & 0.96 & 0.09 \\
&  &  & sink & 0.01 & 0.03 \\
&  &  & bathtub & 0.00 & 0.03 \\

\bottomrule
\end{tabular}
\end{table}

\subsection{Generalization to Unseen Queries}
To evaluate generalization to unseen vocabulary, we separate performance into two categories based on whether the target queries are present in the training data. As shown in Tab.~\ref{tab:generalization}, performance on unseen vocabulary remains overall comparable to performance on seen queries, confirming our method's open-vocabulary generalization. This is attributed to the use of pretrained text embedders that ensure that our models learn mappings from semantically abstracted representations.    

\begin{table}[h]
\vspace{-0.3cm}
\centering
\caption{Performance Breakdown to seen vs. unseen queries.}
\label{tab:generalization}
\begin{tabular}{lcccc}
\toprule
 & \multicolumn{2}{c}{Seen} & \multicolumn{2}{c}{Unseen} \\
\cmidrule(lr){2-3} \cmidrule(lr){4-5}
Benchmark & SR & SPL & SR & SPL \\
\midrule
SymSearch & 0.859 & 0.277 & 0.945 & 0.300 \\
OmniGibson & 0.850 & 0.431 & 0.809 & 0.398 \\
\bottomrule
\end{tabular}
\vspace{-0.3cm}
\end{table}

\subsection{Ablations}
\label{sec:ablations}
In the following, we ablate some design choices and validate the importance of different components introduced in our method. For a fair comparison with LLM-based baselines, we also ablate the choice of the LLM variant.
\subsubsection{Our Method}
We present additional ablations in Tab.~\ref{tab:ablations}, grouped into three categories: 
\paragraph{Design Choices}: We find that the default room exploration score balances the exploration-exploitation behavior.  For instance, a high default room exploration score encourages gathering information across the entire environment, resulting in higher SR but lower SPL. By replacing CLIP embeddings, we further demonstrate that our method is largely agnostic to the choice of embedder, provided that the embedder captures rich semantic information. Finally, the weight controlling the influence of the room score on object scores appears relatively insensitive to its exact value, although assigning too much weight can reduce the overall SR, as all objects in a room will end up with the same score.

\paragraph{Importance of Procedural Data Generation}: Instead of procedurally generating objects as in Eq. 5 (Rooms $\xrightarrow{}$ Categories $\xrightarrow{}$ Objects), we consider two ablations: (i) generating all household objects with a single prompt, and (ii) generating rooms and their corresponding object sets. We observe that introducing intermediate layers increases the diversity of the generated data, thereby improving the final results. The procedural data generation pipeline enables diversity and scale, which directly improve the generalization of learned relational priors and prevent overfitting to the training vocabulary. 
\paragraph{Impact of Data Scale}: Increasing the dataset size progressively improves both SR and SPL. Training on only 10\% or 25\% of the generated data yields performance similar to similarity-based baselines (noisy priors), while increasing the dataset size progressively improves both SR and SPL, approaching the performance of our best performing agent. 

\begin{table}[t]
\centering
\caption{\textbf{Ablation study on SymSearch.}}
\setlength{\tabcolsep}{4pt}
\footnotesize
\begin{tabular}{lccc}
\toprule
\textbf{Agent} & SR $\uparrow$ & SPL $\uparrow$ &  N Steps $\downarrow$ \\
\midrule

\multicolumn{4}{l}{\textbf{Design Choices and Hyperparameters}} \\
\midrule
SCOUT ($\Delta$=0.3)                           & 0.792 & 0.270 & 24 \\
SCOUT (Room Weigth: 0.7)             & 0.814 & 0.277 & 23 \\
SCOUT (Room Weigth: 0.5)             & 0.825 & 0.273 & 22 \\
SCOUT (Room Weigth: 0.1)              & 0.865 & 0.275 & 21 \\
SCOUT (Unknown Room Score: 0.3)  & 0.796 & 0.233 & 28 \\
SCOUT (Unknown Room Score: 0.5)  & 0.834 & 0.275 & 23 \\
SCOUT (Unknown Room Score: 0.9) & 0.871 & 0.206 & 22 \\
SCOUT (CLIP)                                   & 0.865 & 0.279 & 20 \\
\midrule
\multicolumn{4}{l}{\textbf{Procedural Generation}} \\
\midrule
SCOUT (Household $\xrightarrow{}$ Objects)                       & 0.810 & 0.244 & 26 \\
SCOUT (Household $\xrightarrow{}$ Rooms $\xrightarrow{}$ Objects) & 0.791 & 0.248 & 26 \\
\midrule
\multicolumn{4}{l}{\textbf{Data Scale}} \\
\midrule
SCOUT (10\% Data) & 0.728 & 0.169 & 31 \\
SCOUT (25\% Data) & 0.710 & 0.167 & 30 \\
SCOUT (50\% Data) & 0.816 & 0.206 & 24 \\
SCOUT (75\% Data) & 0.865 & 0.222 & 22 \\
\midrule
\multicolumn{4}{l}{\textbf{Full Method}} \\
\midrule
SCOUT & 0.876 & 0.282 & 20 \\
\bottomrule
\end{tabular}
\label{tab:ablations}
\end{table}

\subsubsection{Baselines}
We preselect the LLM model and variant for LLM-based baselines by evaluating the performance of MoMaLLM on 50 sample episodes from \benchmark. Our goal is to choose the best model variant that balances success rate, inference time, and cost. All models were evaluated under similar runtime conditions: local models on an NVIDIA GeForce RTX 4060 Ti, and the others under comparable network settings. The results are summarized in Tab.~\ref{tab:llm_variants}.
Mistral~\cite{Jiang2023Mistral7} and LLaMA~\cite{touvron2023llama} are open-source models that can be run locally and have relatively small sizes (7B and 8B parameters, respectively). However, both perform poorly in this setting. In particular, Mistral frequently fails to follow instructions and often produces invalid outputs. LLaMA is more reliable in producing parseable responses but appears to lack the common-sense reasoning required to perform the task efficiently.
In contrast, the proprietary models from DeepSeek~\cite{guo2025deepseek}, Claude~\cite{anthropic2024claude}, and OpenAI~\cite{achiam2023gpt} consistently exhibit stronger common-sense behavior and achieve higher success rates. While \texttt{deepseek-reasoner} achieves the highest success rate, its additional thinking mode significantly increases inference time, which limits large-scale evaluations. While \texttt{deepseek-chat} runs faster, it suffers from lower performance. Both \texttt{claude-sonnet-4-5} and \texttt{gpt-4o} perform well and are relatively fast, but are significantly more expensive. Overall, we find that \texttt{gpt-5-mini} provides the best trade-off among performance, runtime, and cost, and use it for all LLM-based experiments.

\begin{table*}[t]
\centering
\scriptsize
\setlength{\tabcolsep}{5pt}

\caption{\textbf{Comparison of LLM variants for interactive object search.}
We report model specifications and performance metrics on \benchmark. Prices are reported per 1 million tokens (Input / Output). We report the \besttext{best}, \secondtext{second-best}, and \thirdtext{third-best} results for each metric, respectively.}
\label{tab:llm_variants}

\begin{tabular}{l l c c c c c c c c}

\toprule
\multicolumn{6}{c}{\textbf{Model Card}} & \multicolumn{4}{c}{\textbf{Performance}} \\
\cmidrule(lr){1-6} \cmidrule(lr){7-10}

Model & Version & Pricing (I/O) & Size & Local & Structured &
SR $\uparrow$ & SPL $\uparrow$ & Steps $\downarrow$ & Time (s) $\downarrow$ \\

\midrule

Mistral & \texttt{Mistral-7B-Instruct-v0.1} & Open-source & 7B & \checkmark & $\times$
& 0.24 & 0.149 & 42 & 146 \\

\midrule

LLaMA & \texttt{Meta-Llama-3.1-8B-Instruct} & Open-source & 8B & \checkmark & $\times$
& 0.24 & 0.139 & 39 & 423 \\

\midrule

DeepSeek (chat) & \texttt{deepseek-chat} & \$0.028 / \$0.42 & 671B & $\times$ & $\times$
& 0.78 & 0.222 & 24 & \secondcell 44 \\

DeepSeek (reasoner) & \texttt{deepseek-reasoner} & \$0.028 / \$0.42 & 671B & $\times$ & $\times$
& \bestcell 0.96 & \bestcell 0.356 & \bestcell 15 & 721 \\

\midrule

Claude & \texttt{claude-sonnet-4-5} & \$3 / \$15 & Undisclosed & $\times$ & \checkmark
& \thirdcell 0.82 & 0.216 & 23 & 171 \\

\midrule

GPT-3.5 & \texttt{gpt-3.5-turbo} & \$3 / \$6 & Undisclosed & $\times$ & $\times$
& 0.50 & 0.179 & 33 & \bestcell 23 \\

GPT-4o-mini & \texttt{gpt-4o-mini} & \$0.15 / \$0.60 & Undisclosed & $\times$ & \checkmark
& 0.68 & 0.172 & 27 & 117 \\

GPT-4o & \texttt{gpt-4o} & \$2.5 / \$10 & Undisclosed & $\times$ & \checkmark
& 0.80 & \thirdcell 0.293 & \thirdcell 20 & \thirdcell 116 \\

GPT-5-mini & \texttt{gpt-5-mini} & \$0.25 / \$2 & Undisclosed & $\times$ & \checkmark
& \secondcell 0.88 & \secondcell 0.304 & \secondcell 19 & 278 \\

\bottomrule
\end{tabular}
\end{table*}

\subsection{Real-World Robot Experiments}
\label{sec:real_robot_experiments}
To evaluate the real-world applicability of our method, we deploy it on a mobile manipulator (Toyota HSR) operating in a multi-room apartment environment with a kitchen, office, and living room. Given only a textual object query, the robot must locate the object using only the onboard RGB-D camera and localization as inputs. We follow the same pipeline described in Sec.~\ref{sec:sg_construction}. Additionally, we use the YOLO-World \cite{cheng2024yolo} model for real-time open-vocabulary semantic segmentation, and we add the target object to the list of labels at initialization. To efficiently evaluate reasoning, manipulation actions, i.e., opening containers, are executed using N2M2~\cite{honerkamp23n2m2} with pre-recorded opening motions defined relative to AR markers. More precisely, the robot first navigates to the Voronoi node closest to the object's estimated location. Then, if the associated AR marker is observed, the corresponding N2M2 motion is executed. To this end, we provide a list of affordances for object labels that can be opened or explored (i.e., the robot navigates to the object and examines it). 

\begin{figure}[t]
    \centering
    \subfloat{\includegraphics[width=0.23\linewidth]{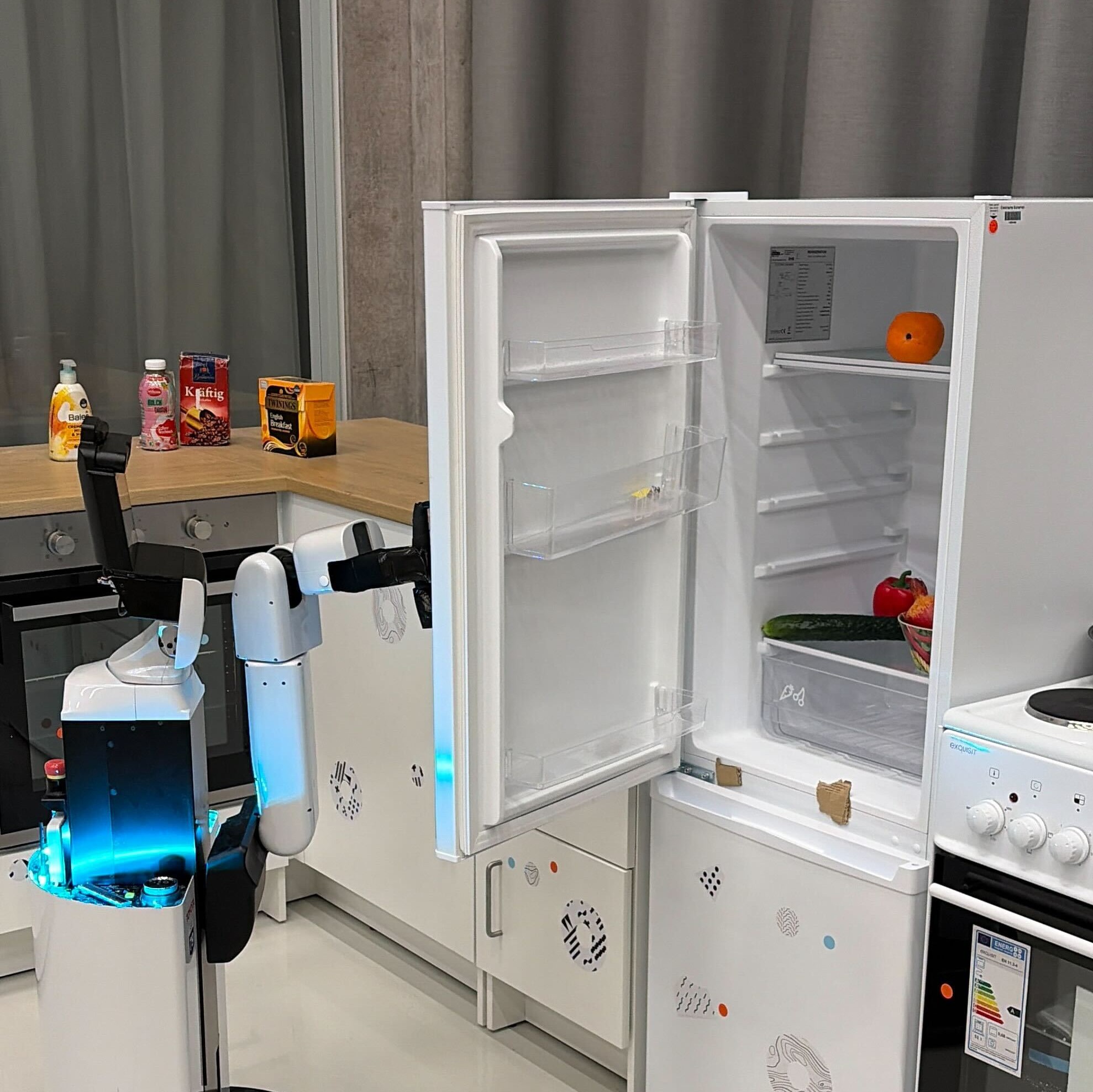}}
    \hfill
    \subfloat{\includegraphics[width=0.23\linewidth]{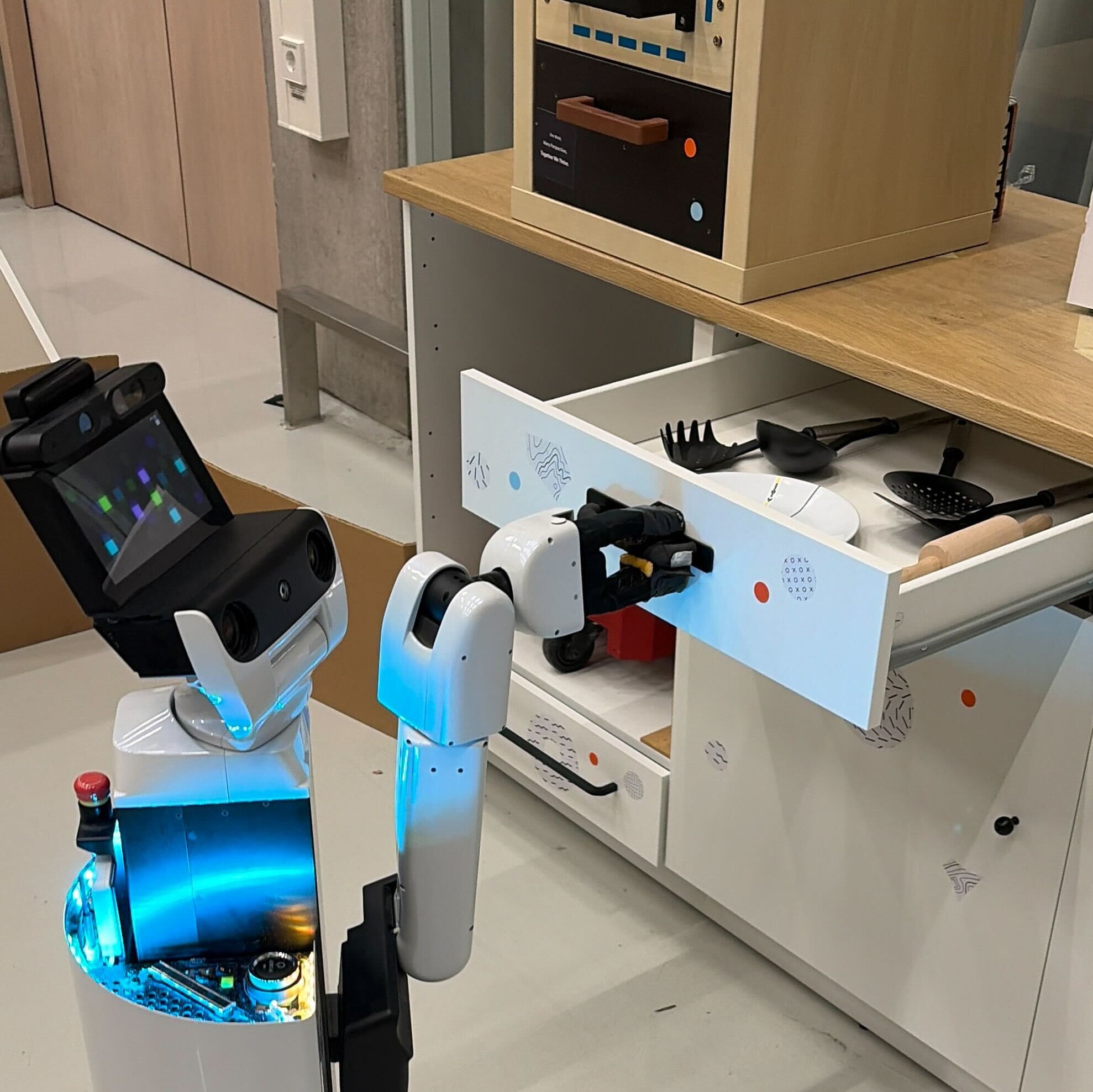}}
    \hfill
    \subfloat{\includegraphics[width=0.23\linewidth]{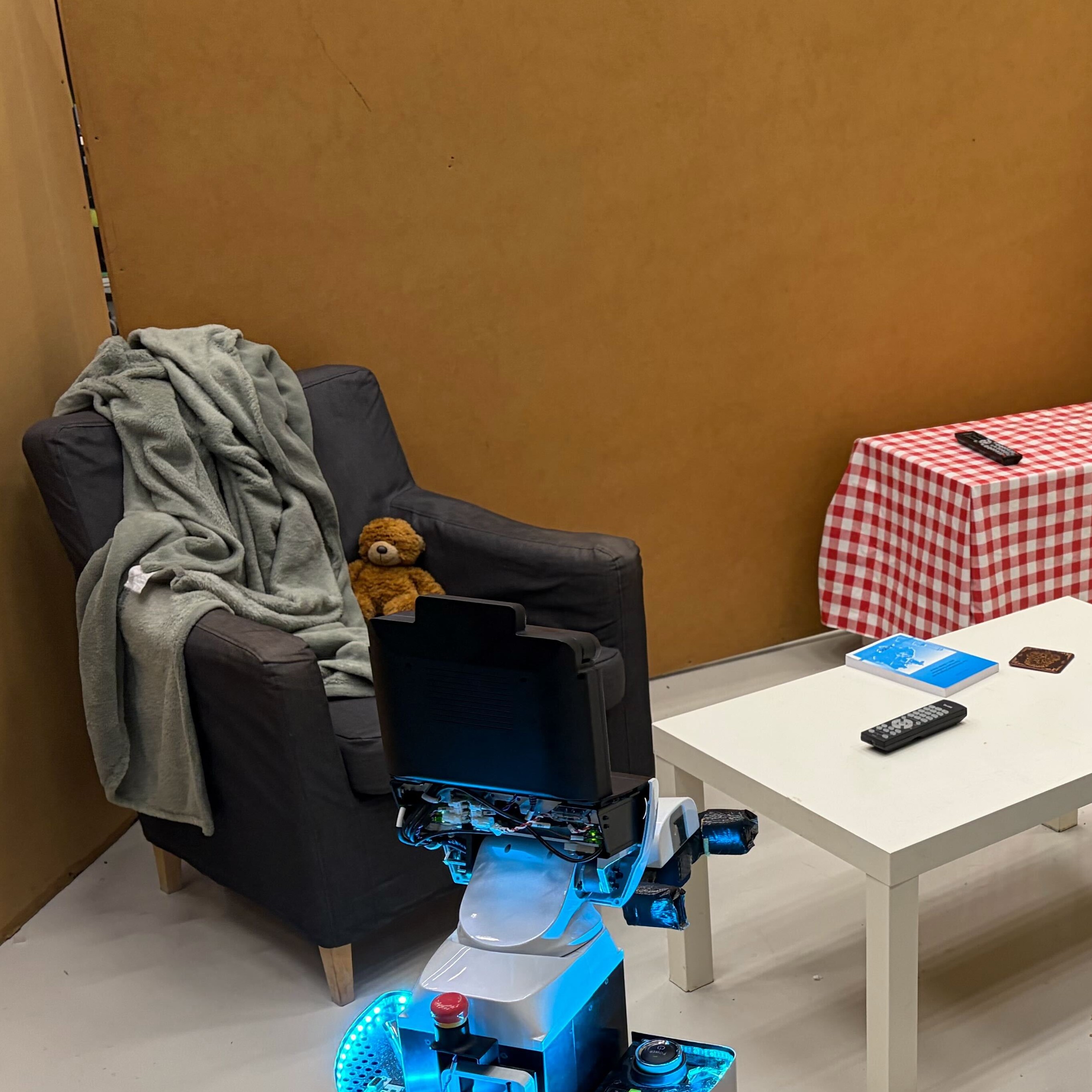}}
    \hfill
    \subfloat{\includegraphics[width=0.23\linewidth]{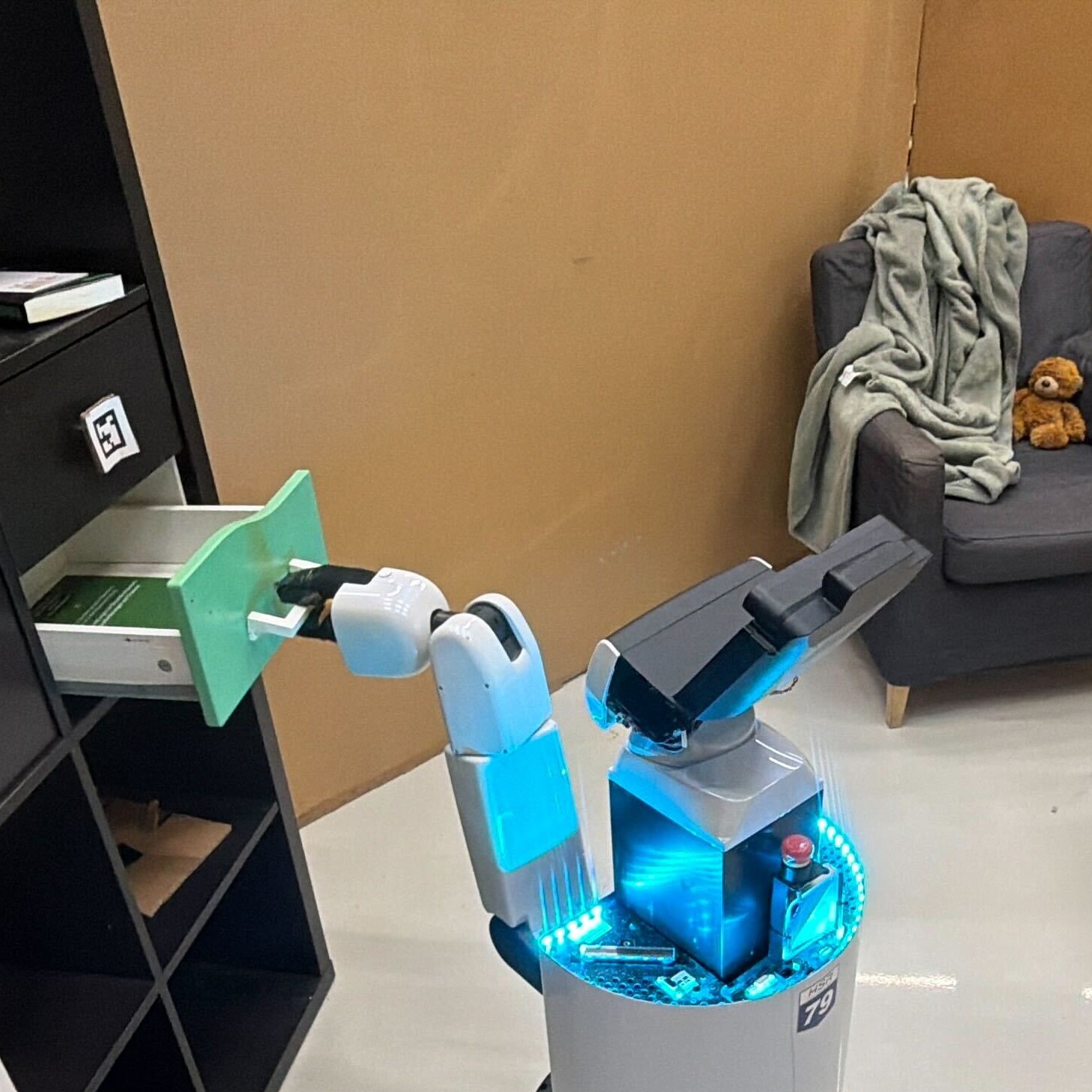}}
    \caption{\textbf{Real-world robot experiments.} The robot's high-level-actions are grounded to the scene through object affordances. The robot can open containers to locate potential target objects (e.g., fridge, drawer, cabinet) or explore on top of furniture (e.g., table, desk, counter).}
    \label{fig:real_robot_examples}
\end{figure}

We conduct 36 real-robot experiments across 12 distinct query object categories, with 7 objects placed inside containers (e.g., fridges or drawers). In all trials, the agent starts from a random location in the environment. We detail the experimental setup and provide examples in the supplementary material. We deem a trial successful if the agent locates the target object while efficiently prioritizing relevant rooms and avoiding unnecessary exploration. 
We categorize failures into three modes based on their root cause: perception, manipulation, and reasoning. Perception failures include missing the target object or misclassifying objects, which can affect room classification. Manipulation failures occur when the agent attempts to interact with a container but does not fully or correctly execute the action (e.g., partially opening a fridge without revealing the target). Reasoning failures arise from our method itself, such as assigning an overly high utility score to an irrelevant room or container, or underestimating a plausible placement. 
In Tab.~\ref{tab:real_world_experiments_details}, we provide descriptions of each evaluation setup. 

\begin{table}[t]
\centering
\caption{\textbf{Results on real world experiments.}
We report outcomes across interactive and non-interactive object search tasks. 
Letters denote result types: S = Success, R = Reasoning Failure, P = Perception Failure, M = Manipulation Failure.}
\label{tab:real_world_experiments_details}
\setlength{\tabcolsep}{3pt}

\begin{tabular}{l c c c c}
\hline
\textbf{Query} & \textbf{Interactive} & \textbf{Location} & \textbf{Room} & \textbf{Outcome} \\
\hline

\multirow{3}{*}{milk bottle} 
& \multirow{3}{*}{\checkmark} 
& \multirow{3}{*}{inside fridge} 
& \multirow{3}{*}{kitchen} & S \\
& & & & S \\
& & & & P \\
\hline

\multirow{3}{*}{remote-controller} 
& \multirow{3}{*}{$\times$} 
& \multirow{3}{*}{on top of table} 
& \multirow{3}{*}{living room} & S \\
& & & & S \\
& & & & P \\
\hline

\multirow{3}{*}{book} 
& \multirow{3}{*}{\checkmark} 
& \multirow{3}{*}{inside shelf} 
& \multirow{3}{*}{living room} & S \\
& & & & S \\
& & & & M \\
\hline

\multirow{3}{*}{blanket} 
& \multirow{3}{*}{$\times$} 
& \multirow{3}{*}{on top of sofa} 
& \multirow{3}{*}{living room} & S \\
& & & & S \\
& & & & S \\
\hline

\multirow{3}{*}{spatula} 
& \multirow{3}{*}{\checkmark} 
& \multirow{3}{*}{inside drawer} 
& \multirow{3}{*}{kitchen} & S \\
& & & & S \\
& & & & S \\
\hline

\multirow{3}{*}{plate} 
& \multirow{3}{*}{\checkmark} 
& \multirow{3}{*}{inside cabinet} 
& \multirow{3}{*}{kitchen} & S \\
& & & & M \\
& & & & P \\
\hline

\multirow{3}{*}{keyboard} 
& \multirow{3}{*}{$\times$} 
& \multirow{3}{*}{on desk} 
& \multirow{3}{*}{office} & S \\
& & & & S \\
& & & & S \\
\hline

\multirow{3}{*}{teddy-bear} 
& \multirow{3}{*}{$\times$} 
& \multirow{3}{*}{on sofa} 
& \multirow{3}{*}{living room} & S \\
& & & & S \\
& & & & P \\
\hline

\multirow{3}{*}{pen} 
& \multirow{3}{*}{\checkmark} 
& \multirow{3}{*}{inside fridge} 
& \multirow{3}{*}{kitchen} & S \\
& & & & S \\
& & & & P \\
\hline

\multirow{3}{*}{fruit} 
& \multirow{3}{*}{\checkmark} 
& \multirow{3}{*}{inside fridge} 
& \multirow{3}{*}{kitchen} & S \\
& & & & S \\
& & & & R \\
\hline

\multirow{3}{*}{cup} 
& \multirow{3}{*}{$\times$} 
& \multirow{3}{*}{inside fridge} 
& \multirow{3}{*}{kitchen} & S \\
& & & & S \\
& & & & P \\
\hline

\multirow{3}{*}{monitor} 
& \multirow{3}{*}{$\times$} 
& \multirow{3}{*}{inside fridge} 
& \multirow{3}{*}{kitchen} & S \\
& & & & P \\
& & & & R \\
\hline

\end{tabular}
\end{table}

\begin{figure*}[t]
    \centering
    \includegraphics[width=\linewidth]{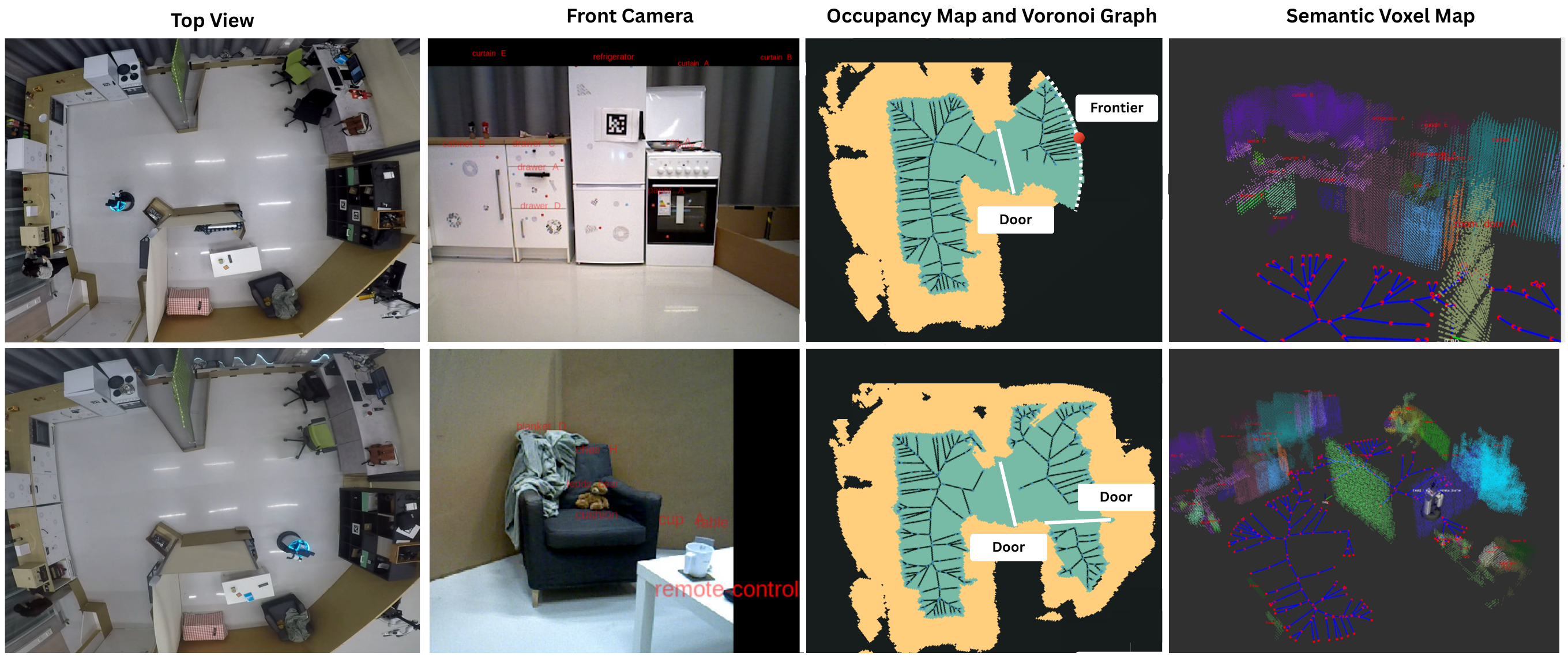}
    \caption{\textbf{Example of a real-world experiment.} The first column shows top-view maps of the environment, which consists of a kitchen, office space, and living room. The second column presents the robot’s front-camera observations with overlaid detected object labels. The third column displays the occupancy map together with the computed Voronoi graph, detected doors, and frontiers. The fourth column shows examples of the 3D semantic voxel maps. \textbf{Top row:} the robot starts in the kitchen and searches for a remote controller located on the living room table. Initially, the agent can observe only the kitchen and part of the office space, which are separated by a frontier. \textbf{Bottom row:} the agent explores the frontier, discovers the living room, and successfully observes the target object.}
    \label{fig:real_experiments}
\end{figure*}

\begin{table}[t]
    \centering
    \footnotesize
    \setlength{\tabcolsep}{4pt}
    \caption{\textbf{Runtime breakdown per timestep.} Average execution time (in seconds) for each component of the pipeline.}
    \label{tab:execution_times}
    \begin{tabular}{cccc}
        \toprule
        \textbf{Scene Graph} &
        \textbf{Inference} &
        \textbf{Execution} &
        \textbf{Total} \\
        \textbf{Construction} &
        \textbf{(Node Selection)} &
        \textbf{(Nav. + Manip.)} &
         \\
        \midrule
        5.036 $\pm$ 0.016 &
        0.211 $\pm$ 0.095 &
        34.348$\pm$ 24.810 &
        39.596 $\pm$ 24.807 \\
        \bottomrule
    \end{tabular}
    \vspace{-0.2cm}
\end{table}

\begin{figure}[t]
    \centering
    \includegraphics[width=\linewidth]{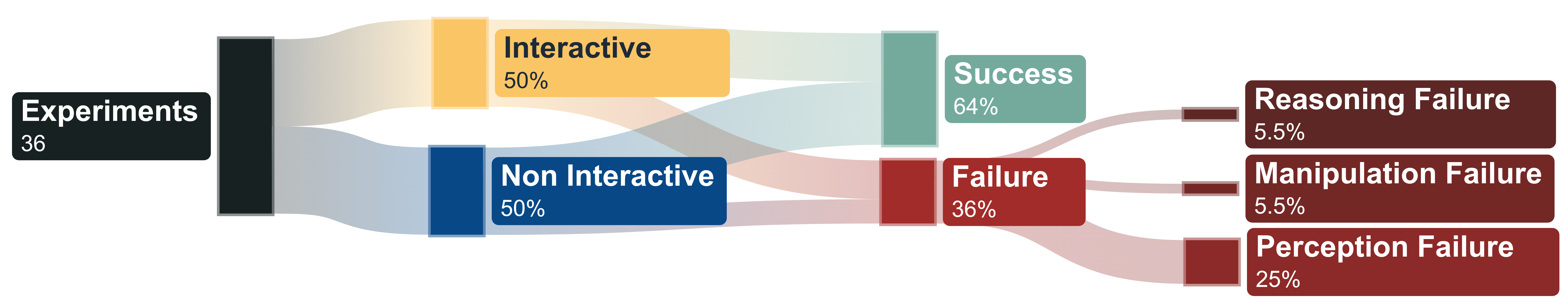}
    \caption{\textbf{Distribution of real-robot experiments.} We conduct a total of 36 experiments, of which 50\% require interactivity with the environment (i.e., the target object is hidden inside containers). Failure modes are primarily due to errors in the perception system.}
    \label{fig:diagram}
    \vspace{-0.3cm}
\end{figure}

Fig.~\ref{fig:diagram} shows the distribution of successful and failed trials, along with the relative frequency of each failure mode. Overall, we achieve a 64\% success rate, including interactive scenarios. For example, when searching for a book, if the robot identifies cabinets in both the kitchen and the living room, it avoids opening kitchen cabinets, reflecting common-sense reasoning about object placement. 
The primary bottleneck is the accuracy of the perception system. Our method relies on a faithful scene graph to operate efficiently; segmentation errors such as duplicated object instances or hallucinated objects can corrupt room classification and mislead the search. In some cases, the robot repeatedly attempts to open visually identical drawers or cabinets that are redundantly represented in the graph, or fails entirely if the target object is not detected. 
From these real-robot experiments, we draw two main conclusions. First, we evaluate the reasoning capabilities of our model under realistic system noise and demonstrate that it reproduces common-sense behaviors distilled from an LLM. Second, we show that our approach can reason over 3D scene graphs, ground this reasoning in the physical world, and extend it to interactive search by mapping object affordances to low-level control policies. 
In Tab.~\ref{tab:execution_times}, we report a breakdown of the per-timestep runtime into three components: scene graph construction, inference time (selecting the next node to explore), and execution time (low-level policy execution, including navigation and control inference). Our method can run efficiently on a real robot without incurring major inference time costs.

\subsection{Discussion and Limitations}
While the rapid advances in modern LLMs make them highly reliable for tasks like the one presented, this reliability comes at the cost of significant computation and constant network access. In contrast, our method runs onboard with low computational overhead while achieving comparable performance, freeing resources for parallel tasks. Moreover, it can serve as a building block in more complex LLM-driven systems, providing efficient object search and reducing the number of expensive queries.
We demonstrate in this paper that our method can: (1) generalize to an open set of object categories across scenes and benchmarks. (2) efficiently leverage the common-sense knowledge in LLMs while maintaining fast runtime inference time by exploiting exploration heuristic and our proposed procedural distillation framework. (3) match and surpass an LLM-planner performance at a fraction of the computational cost. (4) effectively transfer to realistic household settings. 
Despite these strengths, real-robot experiments reveal limitations. Performance depends heavily on the quality of the generated 3D scene graph and the reliability of real-time perception. Errors in object detection, localization, or room classification can lead to suboptimal or inefficient exploration. Moreover, our method encodes typical object–room relationships and implicitly assumes similar household structures. It does not currently model household- or user-specific object placement preferences, which may limit robustness in diverse real-world environments.

\section{Conclusion}
\label{sec:discussion}
We introduced \methodname (\methodfullname), a method for interactive object search that leverages the semantic structure of scene graphs and human-inspired search heuristics. Our results demonstrate that lightweight relational semantic models distilled from LLM knowledge can effectively guide exploration. We demonstrate that \methodname outperforms embedding-based heuristics and matches or exceeds LLM-based planners at a fraction of the computational cost. We further validate the practicality of our approach through successful deployment on a mobile manipulator in a real household environment. We publicly release the code upon acceptance. Future work will address both adapting the learned utility scores to specific households by integrating observations on the fly and generalizing the method to more diverse human-centered environments.

\section*{Acknowledgments}
This work was funded by Toyota Motor Europe.

\bibliographystyle{IEEEtran}
\bibliography{references}

\begin{thebibliography}{10}
\providecommand{\url}[1]{#1}
\csname url@samestyle\endcsname
\providecommand{\newblock}{\relax}
\providecommand{\bibinfo}[2]{#2}
\providecommand{\BIBentrySTDinterwordspacing}{\spaceskip=0pt\relax}
\providecommand{\BIBentryALTinterwordstretchfactor}{4}
\providecommand{\BIBentryALTinterwordspacing}{\spaceskip=\fontdimen2\font plus
\BIBentryALTinterwordstretchfactor\fontdimen3\font minus \fontdimen4\font\relax}
\providecommand{\BIBforeignlanguage}[2]{{%
\expandafter\ifx\csname l@#1\endcsname\relax
\typeout{** WARNING: IEEEtran.bst: No hyphenation pattern has been}%
\typeout{** loaded for the language `#1'. Using the pattern for}%
\typeout{** the default language instead.}%
\else
\language=\csname l@#1\endcsname
\fi
#2}}
\providecommand{\BIBdecl}{\relax}
\BIBdecl

\bibitem{hidalgo2005human}
B.~Hidalgo-Sotelo, A.~Oliva, and A.~Torralba, ``Human learning of contextual priors for object search: where does the time go?'' in \emph{Proc. of the IEEE Conf. Computer Vision Pattern Recognition}, 2005, pp. 86--86.

\bibitem{mack2011object}
S.~C. Mack and M.~P. Eckstein, ``Object co-occurrence serves as a contextual cue to guide and facilitate visual search in a natural viewing environment,'' \emph{Journal of vision}, vol.~11, no.~9, pp. 9--9, 2011.

\bibitem{gadre2023cows}
S.~Y. Gadre, M.~Wortsman, G.~Ilharco, L.~Schmidt, and S.~Song, ``Cows on pasture: Baselines and benchmarks for language-driven zero-shot object navigation,'' in \emph{Proc. of the IEEE Conf. Computer Vision Pattern Recognition}, 2023, pp. 23\,171--23\,181.

\bibitem{yokoyama2024vlfm}
N.~Yokoyama, S.~Ha, D.~Batra, J.~Wang, and B.~Bucher, ``Vlfm: Vision-language frontier maps for zero-shot semantic navigation,'' in \emph{Proc. of the IEEE Int. Conf. on Robotics and Automation}, 2024, pp. 42--48.

\bibitem{bajpai2025uncertainty}
U.~Bajpai, J.~R{\"u}ckin, C.~Stachniss, and M.~Popovi{\'c}, ``Uncertainty-informed active perception for open vocabulary object goal navigation,'' \emph{arXiv preprint arXiv:2506.13367}, 2025.

\bibitem{zhang2025apexnav}
M.~Zhang, Y.~Du, C.~Wu, J.~Zhou, Z.~Qi, J.~Ma, and B.~Zhou, ``Apexnav: An adaptive exploration strategy for zero-shot object navigation with target-centric semantic fusion,'' \emph{arXiv preprint arXiv:2504.14478}, 2025.

\bibitem{shah2023navigation}
D.~Shah, M.~R. Equi, B.~Osi{\'n}ski, F.~Xia, B.~Ichter, and S.~Levine, ``Navigation with large language models: Semantic guesswork as a heuristic for planning,'' in \emph{Conference on Robot Learning}, 2023.

\bibitem{wu2024voronav}
P.~Wu, Y.~Mu, B.~Wu, Y.~Hou, J.~Ma, S.~Zhang, and C.~Liu, ``Voronav: Voronoi-based zero-shot object navigation with large language model,'' \emph{arXiv preprint arXiv:2401.02695}, 2024.

\bibitem{loo2025open}
J.~Loo, Z.~Wu, and D.~Hsu, ``Open scene graphs for open-world object-goal navigation,'' \emph{The International Journal of Robotics Research}, 2025.

\bibitem{yin2024sg}
H.~Yin, X.~Xu, Z.~Wu, J.~Zhou, and J.~Lu, ``Sg-nav: Online 3d scene graph prompting for llm-based zero-shot object navigation,'' \emph{Advances in neural information processing systems}, vol.~37, 2024.

\bibitem{ge2024commonsense}
W.~Ge, C.~Tang, and H.~Zhang, ``Commonsense scene graph-based target localization for object search,'' in \emph{Proc. of the IEEE/RSJ Int. Conf. on Intelligent Robots and Systems}, 2024, pp. 13\,318--13\,325.

\bibitem{zhang2025language}
L.~Zhang, Z.~Li, K.~Cai, Q.~Huang, Z.~Bing, and A.~Knoll, ``Language-enhanced mobile manipulation for efficient object search in indoor environments,'' \emph{arXiv preprint arXiv:2508.20899}, 2025.

\bibitem{menonopen}
R.~Menon, Y.~Schmiede, M.~Bennewitz, and H.~Blum, ``Open-vocabulary and semantic-aware reasoning for search and retrieval of objects in dynamic and concealed spaces,'' \emph{IROS Workshop on Perception and Planning for Mobile Manipulation in Changing Environments}, 2025.

\bibitem{chalvatzaki2023learning}
G.~Chalvatzaki, A.~Younes, D.~Nandha, A.~T. Le, L.~F. Ribeiro, and I.~Gurevych, ``Learning to reason over scene graphs: a case study of finetuning gpt-2 into a robot language model for grounded task planning,'' \emph{Frontiers in Robotics and AI}, vol.~10, p. 1221739, 2023.

\bibitem{liu2404delta}
Y.~Liu, L.~Palmieri, S.~Koch, I.~Georgievski, and M.~Aiello, ``Delta: Decomposed efficient long-term robot task planning using large language models,'' \emph{arXiv preprint arXiv:2404.03275}, 2025.

\bibitem{booker2024embodiedrag}
M.~Booker, G.~Byrd, B.~Kemp, A.~Schmidt, and C.~Rivera, ``Embodiedrag: Dynamic 3d scene graph retrieval for efficient and scalable robot task planning,'' \emph{arXiv preprint arXiv:2410.23968}, 2024.

\bibitem{honerkamp2024language}
D.~Honerkamp, M.~B{\"u}chner, F.~Despinoy, T.~Welschehold, and A.~Valada, ``Language-grounded dynamic scene graphs for interactive object search with mobile manipulation,'' \emph{IEEE Robotics and Automation Letters}, 2024.

\bibitem{mohammadi2025more}
M.~Mohammadi, D.~Honerkamp, M.~B{\"u}chner, M.~Cassinelli, T.~Welschehold, F.~Despinoy, I.~Gilitschenski, and A.~Valada, ``More: Mobile manipulation rearrangement through grounded language reasoning,'' in \emph{Proc. of the IEEE/RSJ Int. Conf. on Intelligent Robots and Systems}, 2025.

\bibitem{parashar2025inference}
S.~Parashar, B.~Olson, S.~Khurana, E.~Li, H.~Ling, J.~Caverlee, and S.~Ji, ``Inference-time computations for llm reasoning and planning: A benchmark and insights,'' \emph{arXiv preprint arXiv:2502.12521}, 2025.

\bibitem{ha2023scaling}
H.~Ha, P.~Florence, and S.~Song, ``Scaling up and distilling down: Language-guided robot skill acquisition,'' in \emph{Conference on Robot Learning}, 2023, pp. 3766--3777.

\bibitem{ravichandran2025distilling}
Z.~Ravichandran, I.~Hounie, F.~Cladera, A.~Ribeiro, G.~J. Pappas, and V.~Kumar, ``Distilling on-device language models for robot planning with minimal human intervention,'' \emph{arXiv preprint arXiv:2506.17486}, 2025.

\bibitem{armeni20193d}
I.~Armeni, Z.-Y. He, J.~Gwak, A.~R. Zamir, M.~Fischer, J.~Malik, and S.~Savarese, ``3d scene graph: A structure for unified semantics, 3d space, and camera,'' in \emph{Proc. of the Int. Conf. Comput. Vis.}, 2019.

\bibitem{werby2024hierarchical}
A.~Werby, C.~Huang, M.~B{\"u}chner, A.~Valada, and W.~Burgard, ``Hierarchical open-vocabulary 3d scene graphs for language-grounded robot navigation,'' in \emph{Proc. of the Robotics: Science and Systems}, 2024.

\bibitem{steinke2025collaborative}
T.~Steinke, M.~B{\"u}chner, N.~V{\"o}disch, and A.~Valada, ``Collaborative dynamic 3d scene graphs for open-vocabulary urban scene understanding,'' in \emph{IEEE/RSJ International Conference on Intelligent Robots and Systems (IROS)}, 2025, pp. 6000--6007.

\bibitem{rana2023sayplan}
K.~Rana, J.~Haviland, S.~Garg, J.~Abou-Chakra, I.~Reid, and N.~Suenderhauf, ``Sayplan: Grounding large language models using 3d scene graphs for scalable robot task planning,'' \emph{arXiv preprint arXiv:2307.06135}, 2023.

\bibitem{schmalstieg2023learning}
F.~Schmalstieg, D.~Honerkamp, T.~Welschehold, and A.~Valada, ``Learning hierarchical interactive multi-object search for mobile manipulation,'' \emph{IEEE Robotics and Automation Letters}, vol.~8, no.~12, pp. 8549--8556, 2023.

\bibitem{schmalstieg2022learning}
------, ``Learning long-horizon robot exploration strategies for multi-object search in continuous action spaces,'' in \emph{The International Symposium of Robotics Research}, 2022, pp. 52--66.

\bibitem{prasanna2024perception}
S.~Prasanna, D.~Honerkamp, K.~Sirohi, T.~Welschehold, W.~Burgard, and A.~Valada, ``Perception matters: Enhancing embodied ai with uncertainty-aware semantic segmentation,'' \emph{arXiv preprint arXiv:2408.02297}, 2024.

\bibitem{chisari2025robotic}
E.~Chisari, J.~O. Von~Hartz, F.~Despinoy, and A.~Valada, ``Robotic task ambiguity resolution via natural language interaction,'' in \emph{IEEE/RSJ International Conference on Intelligent Robots and Systems (IROS)}, 2025, pp. 14\,821--14\,827.

\bibitem{ginting2024seek}
M.~F. Ginting, S.-K. Kim, D.~D. Fan, M.~Palieri, M.~J. Kochenderfer, and A.-a. Agha-Mohammadi, ``Seek: Semantic reasoning for object goal navigation in real world inspection tasks,'' \emph{arXiv preprint arXiv:2405.09822}, 2024.

\bibitem{savva2019habitat}
M.~Savva, A.~Kadian, O.~Maksymets, Y.~Zhao, E.~Wijmans, B.~Jain, J.~Straub, J.~Liu, V.~Koltun, J.~Malik \emph{et~al.}, ``Habitat: A platform for embodied ai research,'' in \emph{Proc. of the IEEE Conf. Computer Vision Pattern Recognition}, 2019, pp. 9339--9347.

\bibitem{szot2021habitat}
A.~Szot, A.~Clegg, E.~Undersander, E.~Wijmans, Y.~Zhao, J.~Turner, N.~Maestre, M.~Mukadam, D.~S. Chaplot, O.~Maksymets \emph{et~al.}, ``Habitat 2.0: Training home assistants to rearrange their habitat,'' \emph{Advances in neural information processing systems}, vol.~34, pp. 251--266, 2021.

\bibitem{yadav2023habitat}
K.~Yadav, R.~Ramrakhya, S.~K. Ramakrishnan, T.~Gervet, J.~Turner, A.~Gokaslan, N.~Maestre, A.~X. Chang, D.~Batra, M.~Savva \emph{et~al.}, ``Habitat-matterport 3d semantics dataset,'' in \emph{Proc. of the IEEE Conf. Computer Vision Pattern Recognition}, 2023, pp. 4927--4936.

\bibitem{yokoyama2024hm3d}
N.~Yokoyama, R.~Ramrakhya, A.~Das, D.~Batra, and S.~Ha, ``Hm3d-ovon: A dataset and benchmark for open-vocabulary object goal navigation. in 2024 ieee,'' in \emph{Proc. of the IEEE/RSJ Int. Conf. on Intelligent Robots and Systems}, 2024, pp. 5543--5550.

\bibitem{kolve2017ai2}
E.~Kolve, R.~Mottaghi, W.~Han, E.~VanderBilt, L.~Weihs, A.~Herrasti, M.~Deitke, K.~Ehsani, D.~Gordon, Y.~Zhu \emph{et~al.}, ``Ai2-thor: An interactive 3d environment for visual ai,'' \emph{arXiv preprint arXiv:1712.05474}, 2017.

\bibitem{li2024behavior}
C.~Li, R.~Zhang, J.~Wong, C.~Gokmen, S.~Srivastava, R.~Mart{\'\i}n-Mart{\'\i}n, C.~Wang, G.~Levine, W.~Ai, B.~Martinez \emph{et~al.}, ``Behavior-1k: A human-centered, embodied ai benchmark with 1,000 everyday activities and realistic simulation,'' \emph{arXiv preprint arXiv:2403.09227}, 2024.

\bibitem{agia2022taskography}
C.~Agia, K.~M. Jatavallabhula, M.~Khodeir, O.~Miksik, V.~Vineet, M.~Mukadam, L.~Paull, and F.~Shkurti, ``Taskography: Evaluating robot task planning over large 3d scene graphs,'' in \emph{Conference on Robot Learning}, 2022, pp. 46--58.

\bibitem{rosinol2021kimera}
A.~Rosinol, A.~Violette, M.~Abate, N.~Hughes, Y.~Chang, J.~Shi, A.~Gupta, and L.~Carlone, ``Kimera: From slam to spatial perception with 3d dynamic scene graphs,'' \emph{The International Journal of Robotics Research}, vol.~40, no. 12-14, pp. 1510--1546, 2021.

\bibitem{hughes2022hydra}
N.~Hughes, Y.~Chang, and L.~Carlone, ``Hydra: A real-time spatial perception system for 3d scene graph construction and optimization,'' \emph{arXiv preprint arXiv:2201.13360}, 2022.

\bibitem{gu2024conceptgraphs}
Q.~Gu, A.~Kuwajerwala, S.~Morin, K.~M. Jatavallabhula, B.~Sen, A.~Agarwal, C.~Rivera, W.~Paul, K.~Ellis, R.~Chellappa \emph{et~al.}, ``Conceptgraphs: Open-vocabulary 3d scene graphs for perception and planning,'' in \emph{Proc. of the IEEE Int. Conf. on Robotics and Automation}, 2024, pp. 5021--5028.

\bibitem{maggio2024clio}
D.~Maggio, Y.~Chang, N.~Hughes, M.~Trang, D.~Griffith, C.~Dougherty, E.~Cristofalo, L.~Schmid, and L.~Carlone, ``Clio: Real-time task-driven open-set 3d scene graphs,'' \emph{IEEE Robotics and Automation Letters}, 2024.

\bibitem{topiwala2018frontier}
A.~Topiwala, P.~Inani, and A.~Kathpal, ``Frontier based exploration for autonomous robot,'' \emph{arXiv preprint arXiv:1806.03581}, 2018.

\bibitem{hart1968formal}
P.~E. Hart, N.~J. Nilsson, and B.~Raphael, ``A formal basis for the heuristic determination of minimum cost paths,'' \emph{IEEE transactions on Systems Science and Cybernetics}, vol.~4, no.~2, pp. 100--107, 1968.

\bibitem{InteriorGS2025}
M.~T.~I. SpatialVerse Research~Team, ``Interiorgs: A 3d gaussian splatting dataset of semantically labeled indoor scenes,'' \url{https://huggingface.co/datasets/spatialverse/InteriorGS}, 2025.

\bibitem{dijkstra2022note}
E.~W. Dijkstra, ``A note on two problems in connexion with graphs,'' in \emph{Edsger Wybe Dijkstra: his life, work, and legacy}, 2022, pp. 287--290.

\bibitem{reimers2019sentence}
N.~Reimers and I.~Gurevych, ``Sentence-bert: Sentence embeddings using siamese bert-networks,'' \emph{arXiv preprint arXiv:1908.10084}, 2019.

\bibitem{radford2021learning}
A.~Radford, J.~W. Kim, C.~Hallacy, A.~Ramesh, G.~Goh, S.~Agarwal, G.~Sastry, A.~Askell, P.~Mishkin, J.~Clark \emph{et~al.}, ``Learning transferable visual models from natural language supervision,'' in \emph{Int. Conf. on Machine Learning}, 2021, pp. 8748--8763.

\bibitem{achiam2023gpt}
J.~Achiam, S.~Adler, S.~Agarwal, L.~Ahmad, I.~Akkaya, F.~L. Aleman, D.~Almeida, J.~Altenschmidt, S.~Altman, S.~Anadkat \emph{et~al.}, ``Gpt-4 technical report,'' \emph{arXiv preprint arXiv:2303.08774}, 2023.

\bibitem{Jiang2023Mistral7}
A.~Q. Jiang, A.~Sablayrolles, A.~Mensch, C.~Bamford, D.~S. Chaplot, D.~de~Las~Casas, F.~Bressand, G.~Lengyel, G.~Lample, L.~Saulnier, L.~R. Lavaud, M.-A. Lachaux, P.~Stock, T.~L. Scao, T.~Lavril, T.~Wang, T.~Lacroix, and W.~E. Sayed, ``Mistral 7b,'' \emph{arXiv preprint arXiv:2310.06825}, 2023.

\bibitem{touvron2023llama}
H.~Touvron, T.~Lavril, G.~Izacard, X.~Martinet, M.-A. Lachaux, T.~Lacroix, B.~Rozi{\`e}re, N.~Goyal, E.~Hambro, F.~Azhar \emph{et~al.}, ``Llama: Open and efficient foundation language models,'' \emph{arXiv preprint arXiv:2302.13971}, 2023.

\bibitem{guo2025deepseek}
D.~Guo, D.~Yang, H.~Zhang, J.~Song, P.~Wang, Q.~Zhu, R.~Xu, R.~Zhang, S.~Ma, X.~Bi \emph{et~al.}, ``Deepseek-r1 incentivizes reasoning in llms through reinforcement learning,'' \emph{Nature}, vol. 645, no. 8081, pp. 633--638, 2025.

\bibitem{anthropic2024claude}
A.~Anthropic, ``The claude 3 model family: Opus, sonnet, haiku,'' \emph{Claude-3 Model Card}, vol.~1, no.~1, p.~4, 2024.

\bibitem{cheng2024yolo}
T.~Cheng, L.~Song, Y.~Ge, W.~Liu, X.~Wang, and Y.~Shan, ``Yolo-world: Real-time open-vocabulary object detection,'' in \emph{Proceedings of the IEEE/CVF conference on computer vision and pattern recognition}, 2024, pp. 16\,901--16\,911.

\bibitem{honerkamp23n2m2}
D.~Honerkamp, T.~Welschehold, and A.~Valada, ``N2m2: Learning navigation for arbitrary mobile manipulation motions in unseen and dynamic environments,'' \emph{IEEE Trans. on Robot.}, vol.~39, no.~5, pp. 3601--3619, 2023.

\end{thebibliography}

\vfill

\end{document}